\documentclass[runningheads]{llncs}

\usepackage[mobile]{eccv}

\usepackage{eccvabbrv}

\usepackage{graphicx}
\usepackage{booktabs}
\usepackage{amsmath}
\usepackage{amssymb}   
\usepackage{mathtools}
\usepackage{amsmath}
\usepackage{enumitem} 
\usepackage{multirow}
\usepackage{bbm} 
\usepackage{makecell}    
\usepackage{amssymb}     
\usepackage{xcolor}      
\usepackage{multicol}
\usepackage{wrapfig}
\usepackage{float}
\usepackage[normalem]{ulem} 
\usepackage{tcolorbox}
\usepackage[pagebackref]{hyperref}
\tcbuselibrary{breakable}

\usepackage[table]{xcolor}
\usepackage{amssymb}
\usepackage{pifont}
\usepackage[normalem]{ulem}
\definecolor{headergray}{RGB}{245,245,245}
\definecolor{groupgray}{RGB}{250,250,250}
\definecolor{myblue}{RGB}{0, 110, 175}  

\definecolor{ourscache}{HTML}{E8F5E9}    
\definecolor{oursnocache}{HTML}{E3F2FD}  

\usepackage[accsupp]{axessibility}  

\usepackage{hyperref}

\usepackage{orcidlink}

\begin{document}

\title{ViBe: Ultra-High-Resolution Video Synthesis Born from Pure Images}

\vspace{-0.2cm}
\titlerunning{ViBe}

\author{
Yunfeng Wu$*$\inst{1,2} \and
Hongying Cheng$*$\inst{1,3} \and
Zihao He\inst{1} \and
Songhua Liu$\dagger$\inst{1}
}
\authorrunning{Wu. et al.}

\institute{
School of Artificial Intelligence, Shanghai Jiao Tong University \and
Xi'an Jiaotong-Liverpool University \and
Jilin University
}

\maketitle

\begin{center}
\vspace{-0.2cm}
    \captionsetup{type=figure}
    \includegraphics[width=0.98\textwidth]{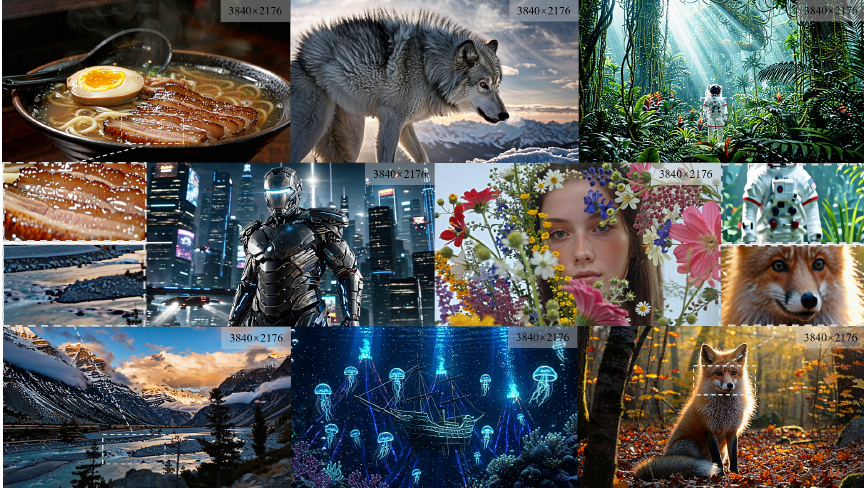}
    \vspace{-0.3cm}
    \captionof{figure}{Ultra-resolution results generated by our Methods built upon Wan2.2~\cite{wan2025wanopenadvancedlargescale}. Resolution is marked on the top-right corner of each result in the format of \texttt{width$\times$height}. Corresponding prompts can be found in the appendix.}
    \vspace{-0.4cm}
    \label{fig:teaser}
\end{center}

\begingroup
\renewcommand\thefootnote{}
\footnotetext{$^{*}$ Equal contribution.}
\footnotetext{$^{\dagger}$ Corresponding author (\texttt{liusonghua@sjtu.edu.cn}).}
\addtocounter{footnote}{-2}
\endgroup

\vspace{-0.2cm}
\begin{abstract}

Transformer-based video diffusion models rely on 3D attention over spatial and temporal tokens, which incurs quadratic time and memory complexity and makes end-to-end training for ultra-high-resolution videos prohibitively expensive. To overcome this bottleneck, we propose a pure image adaptation framework that upgrades a video Diffusion Transformer pre-trained at its native scale to synthesize higher-resolution videos. Unfortunately, naively fine-tuning with high-resolution images alone often introduces noticeable noise due to the image–video modality gap. To address this, we decouple the learning objective to separately handle modality alignment and spatial extrapolation. At the core of our approach is \textit{Relay LoRA}, a two-stage adaptation strategy. In the first stage, the video diffusion model is adapted to the image domain using low-resolution images to bridge the modality gap. In the second stage, the model is further adapted with high-resolution images to acquire spatial extrapolation capability. During inference, only the high-resolution adaptation is retained to preserve the video generation modality while enabling high-resolution video synthesis. To enhance fine-grained detail synthesis, we further propose a \textit{High-Frequency-Awareness-Training-Objective}, which explicitly encourages the model to recover high-frequency components from degraded latent representations via a dedicated reconstruction loss. 
Extensive experiments demonstrate that our method produces ultra-high-resolution videos with rich visual details without requiring any video training data, even outperforming previous state-of-the-art models trained on high-resolution videos by $0.8$ on the VBench benchmark. Codes will be available \href{https://github.com/WillWu111/ViBe}{here}.

\end{abstract}

\section{Introduction}
\label{sec:intro}

Empowered by the attention mechanism’s ability to capture complex token-wise dependencies~\cite{vaswani2017attention}, Diffusion Transformers (DiTs)~\cite{peebles2023scalablediffusionmodelstransformers} have shown remarkable performance in generating high-quality images~\cite{chen2024pixartsigmaweaktostrongtrainingdiffusion, esser2024scalingrectifiedflowtransformers,gao2024luminat2xtransformingtextmodality} and videos~\cite{kong2025hunyuanvideosystematicframeworklarge, hacohen2024ltxvideorealtimevideolatent, yang2025cogvideoxtexttovideodiffusionmodels,hong2022cogvideolargescalepretrainingtexttovideo}. 
However, the quadratic complexity of attention with respect to spatial resolution makes scaling up these models computationally prohibitive, and this issue becomes even more severe in video generation, where an additional temporal dimension further amplifies the cost. 
As a result, most existing diffusion models are trained at relatively low resolutions.
This stands in stark contrast to the rapidly growing demand for ultra-high-definition content, such as 4K videos, in modern visual applications.

\begin{wrapfigure}{r}{0.58\textwidth} 
    \vspace{-0.9cm}
    \centering
    \includegraphics[width=1\linewidth]{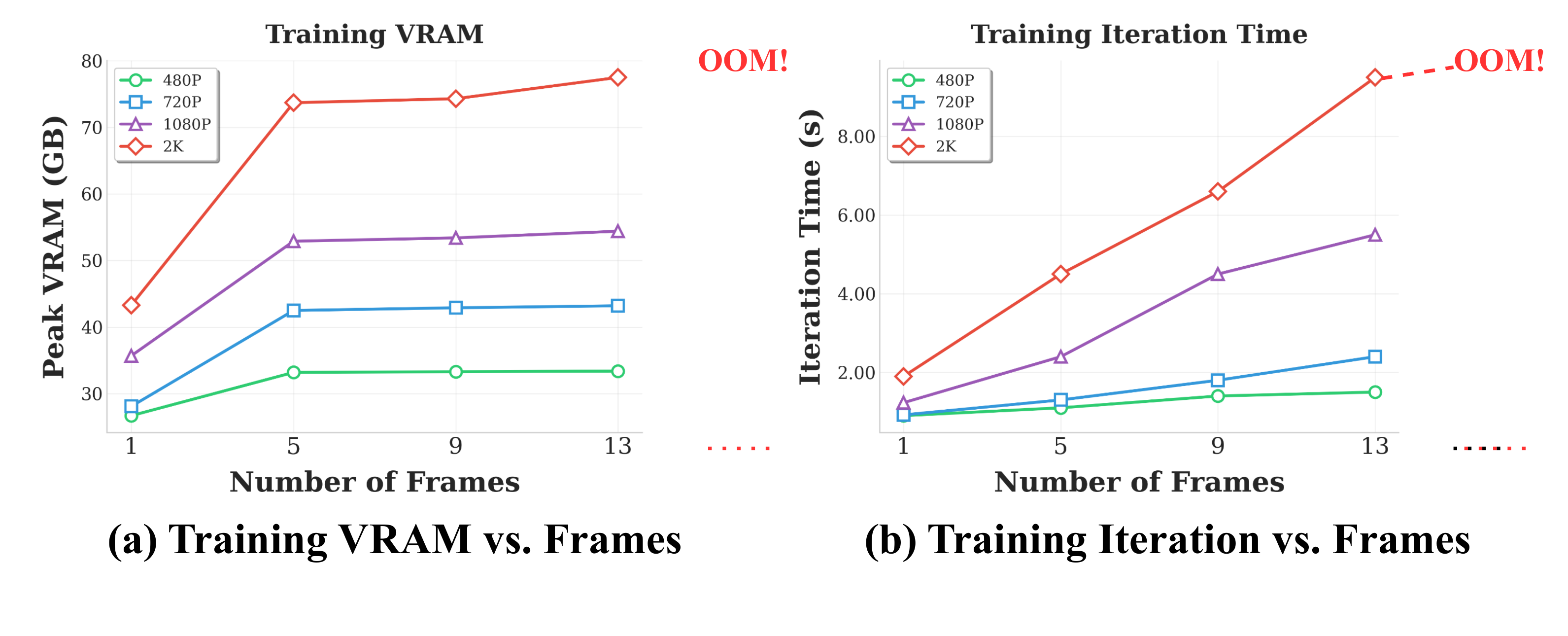}
    \vspace{-0.9cm}
    \caption{The two plots show how training VRAM consumption and iteration time scale with the number of frames under different resolutions.}
    \vspace{-0.9cm}
    \label{fig:time}
\end{wrapfigure}

We report how VRAM usage and iteration time scale with the number of frames under different resolutions to provide an intuitive comparison between training with high-resolution images and videos. As the resolution increases, VRAM consumption rises sharply, as shown in Fig.~\ref{fig:time}(a); as the number of frames grows, the iteration time increases approximately linearly, as shown in Fig.~\ref{fig:time}(b). Considering the substantial challenges of training high-resolution video generation models, we pose a natural question in this work: \emph{Can we generate ultra-high-resolution videos using existing video generators pre-trained only at lower resolution, such as 480P, by training solely with image data?}

\begin{figure*}[!t]
    \centering
    \includegraphics[width=\linewidth]{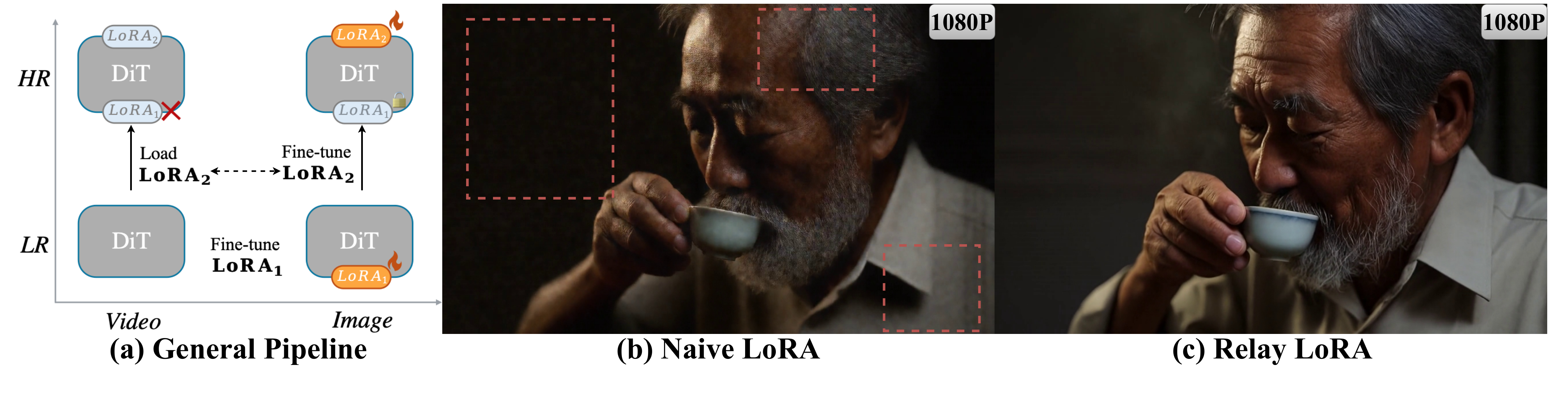}
    \vspace{-0.6cm}
    \caption{The plots on the left shows the general pipeline of our Relay LoRA. The two panels on the right provide a qualitative analysis on Wan2.2~\cite{wan2025wanopenadvancedlargescale}: naive high-resolution image fine-tuning (Naive LoRA) introduces noticeable artifacts, whereas ours methods (Relay LoRA) effectively removes them. 
    }
    \vspace{-0.4cm}
    \label{fig:motivation2}
\end{figure*}

However, this is a highly non-trivial problem--direct training on images while conducting video inference can yield noticeable noise, as shown in Fig.~\ref{fig:motivation2}(b) and (c), which we attribute to the gap between the image and video modalities. 
This observation highlights the need for a principled approach that can bridge the modality gap while enabling spatial extrapolation from image data.

In this paper, we reveal that the core to alleviate this drawback lies decoupling the learning objective to retain only the benefits of spatial extrapolation while mitigating the adverse effects caused by the modality gap between images and videos. 
We thus propose a two-stage adaptation strategy termed \emph{Relay LoRA}~\cite{liu2024fictitioussyntheticdataimprove}, which explicitly separate the two adaptation processes: modality alignment and spatial extrapolation, as shown in Fig.~\ref{fig:motivation2}(a). Specifically, in Stage~1, we fine-tune a base video DiT on \emph{low-resolution images} with LoRA~\cite{hu2021loralowrankadaptationlarge} and obtain $\mathrm{LoRA}_1$, which familiarizes the video model with the image modality. 
Building upon $\mathrm{LoRA}_1$, Stage~2 fine-tunes the model on \emph{high-resolution images} to obtain another LoRA, termed $\mathrm{LoRA}_2$, which equips the model with spatial extrapolation capability. 
Since inference is conducted for \emph{high-resolution videos} rather than images, only $\mathrm{LoRA}_2$—the spatial extrapolation adapter—is attached to the base DiT, while $\mathrm{LoRA}_1$—the video-to-image adapter—is discarded.

Moreover, to enhance fine-grained detail rendering, we introduce a \textit{High-Frequency-Awareness-Training-Objective (HFATO)}. HFATO explicitly encourages the model to recover high-frequency components from degraded latent representations, thereby improving its ability to reconstruct subtle textures and sharp structural patterns. To better exploit the degraded latents during training, we couple the latent degradation module with a dedicated reconstruction loss. This additional supervision regularizes the optimization toward high-frequency preservation. Empirically, we observe that this design further improves fine-grained detail fidelity and structural clarity, particularly in challenging high-resolution scenarios.

We conduct extensive experiments on modern DiT-based video generation models, \textit{e.g.}, Wan2.2~\cite{wan2025wanopenadvancedlargescale}. As shown in Fig.~\ref{fig:teaser}, our method synthesizes 4K-resolution videos with strong semantic coherence and visually compelling fine-grained details. Quantitatively, it achieves state-of-the-art results on VBench~\cite{huang2023vbenchcomprehensivebenchmarksuite}, even surpassing prior training-based methods that are trained on real high-resolution video data in both generation quality and consistency. Our contributions are summarized as follows:

\begin{itemize}
\item Conceptually, we delve into the field of ultra-high-resolution video generation and, to the best of our knowledge, present the first method tailored to DiT architectures such as Wan2.2~\cite{wan2025wanopenadvancedlargescale} that achieves ultra-high-resolution video generation using pure image data for training.

\item Technically, we propose a Relay LoRA strategy that mitigates the noise induced by image-only training, augmented with a high frequency awareness training objective to strengthen fine-detail processing.

\item Experimentally, extensive experiments and user evaluations confirm that our method achieves superior performance in ultra-high-resolution video generation, outperforming training-based counterparts.
\end{itemize}

\vspace{-0.1cm}

\section{Related Work}
\subsection{Diffusion Models for Video Generation}
Diffusion models (DMs)~\cite{sohl2015deep,ho2020denoising} have been widely studied for video generation and have attracted substantial attention. In contrast to conventional UNet-based video diffusion models~\cite{ho2022videodiffusionmodels,he2023latentvideodiffusionmodels,chen2023videocrafter1opendiffusionmodels,blattmann2023stablevideodiffusionscaling}, which rely on convolutional backbones, Diffusion Transformer (DiT)~\cite{peebles2023scalablediffusionmodelstransformers} uses a Transformer architecture as the primary denoising backbone. This architectural shift allows DiT to model long-range dependencies more effectively and to represent more complex spatiotemporal relationships in video sequences. Early studies~\cite{yang2025cogvideoxtexttovideodiffusionmodels, jin2025pyramidalflowmatchingefficient} demonstrated the effectiveness of spatiotemporal Transformers that employ self-attention with a global receptive field. More recently, LTX-Video~\cite{hacohen2024ltxvideorealtimevideolatent} improved the coupling between the Video-VAE and the denoising Transformer, and Wan2.1~\cite{wan2025wanopenadvancedlargescale} and Hunyuan~\cite{kong2025hunyuanvideosystematicframeworklarge}, both trained on large-scale video datasets, achieved strong performance in realistic video generation. Despite the substantial gains in generation quality, the native resolution of these models remains limited for high-quality applications.

\subsection{High-Resolution Visual Generation}
In the image domain, training-free high-resolution visual generation has been widely studied. Most existing approaches~\cite{podell2023sdxlimprovinglatentdiffusion} are built on U-Net~\cite{ronneberger2015unetconvolutionalnetworksbiomedical} and often suffer from repetitive patterns in high-resolution synthesis due to limited local receptive fields. Several works~\cite{he2023scalecraftertuningfreehigherresolutionvisual,bartal2023multidiffusionfusingdiffusionpaths,du2023demofusiondemocratisinghighresolutionimage} enlarge receptive fields through dilated convolutions and fuse local and global patches to mitigate repetition. With the emergence of recent foundational diffusion models~\cite{labs2025flux1kontextflowmatching}, DiT has become the dominant architecture, benefiting from the capacity of attention mechanisms to model complex token-wise dependencies. Training-free methods~\cite{bu2025hiflowtrainingfreehighresolutionimage,du2024imaxmaximizeresolutionpotential} built on DiT can preserve global layout consistency while synthesizing fine-grained visual details.

In the video domain, high-resolution synthesis with DiT-based models introduces additional challenges, including substantial computational overhead, blur, and structural distortion. Some works~\cite{ye2025supergenefficientultrahighresolutionvideo} focus on the system infrastructure required for efficient high-resolution video generation, but the generation quality remains constrained. Most current video generation methods~\cite{ren2025turbo2k,hu2025ultragenhighresolutionvideogeneration,qiu2025cinescalefreelunchhighresolution} rely on fine-tuning with high-resolution data, which imposes prohibitive computational demands. In this work, we propose a fully training-based approach that uses pure image data tailored for modern DiT architectures, while ensuring global layout consistency and producing fine-grained details in high-resolution synthesis.

\vspace{-0.1cm}

\section{Methods}
Generating 4K videos is challenging because it requires both fine-grained detail synthesis and coherent high-level semantic structure under a tight compute budget. 
Following prior work~\cite{meng2021sdedit}, we adopt a coarse-to-fine pipeline: we first generate a video at the model's native resolution from the text prompt to establish global layout, object identities, and motion semantics. We then produce a high resolution version based on the initial output, so that the second stage focuses on recovering high-frequency details while preserving the coarse structure.

\begin{figure*}[!t]
    \centering
    \includegraphics[width=\linewidth]{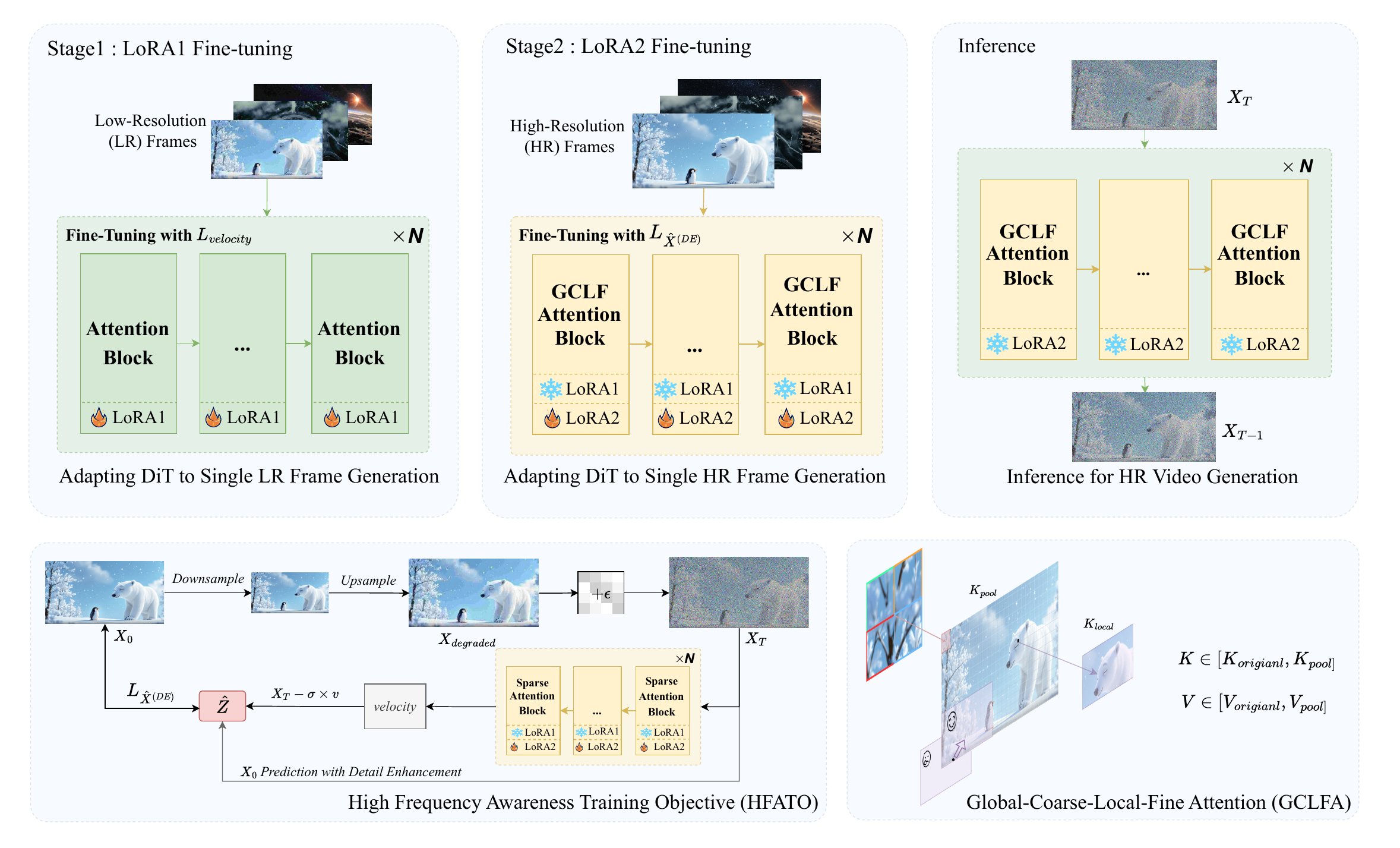}
    \vspace{-0.4cm}
    \caption{\textbf{Upper Left}: Stage-1 fine-tuning trains $\mathrm{LoRA}_1$ to adapt the DiT backbone to single low-resolution frame generation. \textbf{Upper Middle}: Stage-2 first merges $\mathrm{LoRA}_1$ into the base model and freezes the merged weights, then trains a Relay $\mathrm{LoRA}_2$ to adapt the DiT backbone to single high-resolution frame generation. \textbf{Upper Right}: During inference, only $\mathrm{LoRA}_2$ is loaded on the base model for video generation. \textbf{Bottom Left}: Global-Coarse-Local-Fine-Attention combines an inward sliding-window local attention with pooled coarse attention. \textbf{Bottom Right}: The training objective first degrades the latents via a downsample--upsample operation, then add noise on the degraded sample. The model applies the predicted flow and computes the loss against the clean latents.}
    \vspace{-0.4cm}
    \label{fig:method}
\end{figure*}

We implement this pipeline within a flow-matching framework that specifies both the training objective and the sampling pipelines (Sec.~\ref{sec:Flow Matching}). 
On top of this backbone, we introduce three key components.
(i) \textbf{Relay LoRA} (Sec.~\ref{sec:Relay LoRA}) decomposes image-based fine-tuning into two stages to mitigate modality-induced artifacts.
(ii) \textbf{Global-Coarse-Local-Fine-Attention (GCLFA)} (Sec.~\ref{sec:Global-Coarse-Local-Fine Attention}) balances global semantic consistency and local detail modeling, enabling effective long-range interactions at high resolution.
(iii) \textbf{High-Frequency-Awareness-Training-Objective (HFATO)} (Sec.~\ref{sec:High Frequency awareness training objective}) enhances high-frequency detail synthesis by explicitly training the model to reconstruct clean latents from degraded latents. Together, these designs improve 4K detail fidelity while maintaining coherent semantics and fine-grained details. An overview of the method is shown in Fig.~\ref{fig:method}.

\subsection{Flow Matching in Video Generation}\label{sec:Flow Matching}
Flow matching~\cite{lipman2023flowmatchinggenerativemodeling} provides a convenient parameterization for diffusion-based video generation by learning a time-dependent velocity field whose integral curve transports a simple base distribution to the data distribution. Given a clean sample $x_0 \sim p_{\text{data}}$ and Gaussian noise $\epsilon \sim \mathcal{N}(0,I)$, flow matching constructs a continuous interpolation
\begin{equation}
    x_t \coloneqq \alpha_t x_0 + \sigma_t \epsilon,\quad t\in[0,1],
\end{equation}
which smoothly bridges the data at $t=0$ and a noise distribution at $t=1$. The model $v_{\theta}$ is trained to approximate the ground-truth velocity along this trajectory by minimizing the weighted mean squared error
\begin{equation}
    \mathcal{L}_{\text{FM}} = \mathbb{E}_{x_{0},\epsilon,t}\!\left[w(t)\left\lVert v_{\theta}(x_{t},t)-v_{t}\right\rVert^{2}\right],
\end{equation}
where $t \sim \mathcal{U}(0,1)$ and $w(t)$ is a time-dependent reweighting that balances learning across different noise levels.
Under the linear noise schedule~\cite{liu2022flowstraightfastlearning} defined by $\alpha_{t}=1-t$, $\sigma_{t}=t$, and $T=1$, we have
\begin{equation}
    \frac{\mathrm{d}x_t}{\mathrm{d}t} = \dot{\alpha}_t x_0 + \dot{\sigma}_t \epsilon = (-1)\,x_0 + (1)\,\epsilon,
\end{equation}
thus the target velocity satisfies
\[
v_{t}\coloneqq \frac{\mathrm{d}x_{t}}{\mathrm{d}t}=\epsilon-x_{0}.
\]
After training, sampling is performed by integrating the probability flow ordinary differential equation (PF-ODE)~\cite{song2021scorebasedgenerativemodelingstochastic},
\begin{equation}
\mathrm{d}x_{t}=v_{\theta}(x_{t},t)\,\mathrm{d}t,\quad x_{T}\sim\mathcal{N}(0,I),\quad t: T\rightarrow 0,
\label{eq:pf_ode}
\end{equation}
which deterministically maps a noise sample to a data sample. In practice, \eqref{eq:pf_ode} is solved with a numerical ODE solver (e.g., Euler or Heun) using a small number of discretization steps.

\subsection{Relay LoRA}\label{sec:Relay LoRA}
Nevertheless, in video generation, a model must capture complex 3D interactions across both spatial and temporal dimensions, whereas image generation involves only spatial structure. As shown in Fig.~\ref{fig:motivation2}(b) and (c), naively fine-tuning the model on high-resolution images and directly using the resulting LoRA~\cite{hu2021loralowrankadaptationlarge} for video inference introduces noticeable artifacts due to the modality gap between images and videos, which can severely degrade visual quality. 

To address this problem, motivated by this observation, we speculate that introducing an auxiliary adaptation mechanism to mitigate the modality gap before fine-tuning for new capabilities, i.e., high-resolution generation, can effectively address this issue. To this end, the naive high-resolution image fine-tuning procedure is decomposed into two stages, denoted as Relay LoRA. In the first stage, a LoRA~\cite{hu2021loralowrankadaptationlarge} module is trained on low-resolution images used for video pretraining, adapting DiTs to single low-resolution frame generation, denoted as $\mathrm{LoRA}_1$. The learned $\mathrm{LoRA}_1$ is merged with the base model to produce a new set of weights,

\begin{equation}
    W^{(1)} = W^{(0)} + \Delta W_{\mathrm{LoRA}_1}, 
    \qquad 
    \Delta W_{\mathrm{LoRA}_1} = \frac{\alpha}{r} B_1 A_1 ,
\end{equation}
where $W^{(0)}$ denotes the frozen base-model weights and $(A_1,B_1)$ are the low-rank factors learned in the first stage, $r$ is the LoRA rank, and $\alpha$ is a scaling hyperparameter, which may still introduce image-induced artifacts. In the second stage, high-resolution images are used to train a second LoRA~\cite{hu2021loralowrankadaptationlarge} module, denoted as $\mathrm{LoRA}_2$, on top of the merged weights $W^{(1)}$, adapting DiTs to single high-resolution frame generation. This two-stage design effectively mitigates modality-induced artifacts. During inference, we load only $\mathrm{LoRA}_2$ into the base model, which is decoupled and learns only the ability to handle spatial high-resolution frame generation. Concretely, for each LoRA-injected weight matrix, we apply
\begin{equation}
    W_{\text{infer}} = W^{(0)} + \Delta W_{\mathrm{LoRA}_2},
    \qquad
    \Delta W_{\mathrm{LoRA}_2} = \frac{\alpha}{r} B_2 A_2 ,
\end{equation}
where $W^{(0)}$ denotes the base-model weights and $(A_2,B_2)$ are the low-rank factors learned in the second stage, $r$ is the LoRA rank, and $\alpha$ is a scaling hyperparameter. Based on our observation, we can design any customized attention variants in the second stage for visual enhancement. Therefore, to further strengthen fine-detail synthesis at high resolution, the method incorporates the GCLFA described in Sec.~\ref{sec:Global-Coarse-Local-Fine Attention}.

\subsection{Global-Coarse-Local-Fine-Attention}\label{sec:Global-Coarse-Local-Fine Attention}
Inspired by FreeSwim~\cite{wu2025freeswimrevisitingslidingwindowattention}, 
which demonstrates that local attention significantly improves fine-grained details, 
we introduce a \textit{Global-Coarse-Local-Fine-Attention (GCLFA)} in Stage~2, 
as illustrated in Fig.~\ref{fig:method} (Bottom Left). 
The local branch enhances high-frequency details, while the global branch preserves semantic coherence.
\paragraph{Local Fine Branch.}
In the fine local branch, naive sliding-window attention truncates the attention window for boundary tokens due to limited spatial context, resulting in an inconsistent number of interactable key/value tokens across queries. 
To ensure a uniform receptive field for all tokens, including those near boundaries, we propose an inward sliding-window attention mechanism. 
Specifically, when a query's attention window extends beyond the video boundary, the window is shifted inward according to the token's spatial location (e.g., using a 480P window for a 480P model). 
This design preserves the receptive-field size used during training and eliminates boundary truncation, which can be formulated as:
\begin{equation}
\label{eq:local}
\begin{aligned}
\Delta^{(q)}_w &= \max\!\left(\frac{w}{2}-x_q,\frac{w}{2}+x_q-W+1,0\right),\\
\Delta^{(q)}_h &= \max\!\left(\frac{h}{2}-y_q,\frac{h}{2}+y_q-H+1,0\right),\\
M_{qk} &=
\begin{cases}
1, & \text{if } \left| x_q - x_k \right| \leq \frac{w}{2} + \Delta^{(q)}_w \\
   & \text{ and } \left| y_q - y_k \right| \leq \frac{h}{2} + \Delta^{(q)}_h, \\
0, & \text{otherwise},
\end{cases}\\
\end{aligned}
\end{equation}

\paragraph{Global Coarse Branch.}
For the global branch, we first apply RoPE~\cite{su2023roformerenhancedtransformerrotary} to $Q$ and $K$ to inject positional information. 
The key and value are then pooled to produce downsampled coarse tokens:
\begin{equation}
\bar{K} = \mathrm{Pool}(K), \quad 
\bar{V} = \mathrm{Pool}(V).
\end{equation}
These coarse tokens provide a compact global representation aligned with the global branch. 
We concatenate them with the original key--value sequence:
\begin{equation}
\tilde{K} = \mathrm{Concat}(K,\bar{K}), \quad 
\tilde{V} = \mathrm{Concat}(V,\bar{V}),
\end{equation}
so that each query attends to both fine local tokens and coarse global tokens:
\begin{equation}
\label{eq:global}
O = \mathrm{Softmax}\!\left(\frac{Q\tilde{K}^{\top}\cdot M}{\sqrt{d}}\right)\tilde{V}.
\end{equation}

Here, $x_{*}$ and $y_{*}$ denote spatial token coordinates; 
$W$ and $H$ are the target spatial dimensions, while $w$ and $h$ are the native window sizes; 
$Q$, $K$, $V$, and $M$ denote the query, key, value, and attention mask matrices; 
$d$ is the feature dimension; 
and $O$ is the attention output. 
We replace the original 3D full attention with this scheme in all self-attention layers, 
which serve as the primary token interaction modules in modern DiT-based video generators such as Wan2.2~\cite{wan2025wanopenadvancedlargescale}.

\subsection{High-Frequency-Awareness-Training-Objective}
\label{sec:High Frequency awareness training objective}

To enhance high-frequency detail synthesis, we propose a \textit{High-Frequency-Awarenss-Training Objective (HFATO)} that explicitly enforces clean-latent reconstruction from degraded noisy inputs. Unlike standard flow-matching training~\cite{labs2025flux1kontextflowmatching}, which perturbs the clean latent with noise and learns to predict the flow at timestep $t$, we first apply a downsample--upsample degradation operator to corrupt high-frequency components before injecting Gaussian noise. 

Under this formulation, the model predicts the flow from latents that are first degraded and then corrupted by noise, rather than from clean latents directly perturbed by noise. This shift in training dynamics encourages the model to operate under high-frequency-deficient conditions and explicitly learn detail restoration. To promote high-frequency generation, we introduce a dedicated reconstruction objective that supervises the recovery of the original clean latent. Specifically, the model estimates the flow on degraded and noisy latents and reconstructs the clean latent accordingly. We minimize a reconstruction loss to enforce consistency between the restored latent and the ground-truth clean latent. Empirically, we find that directly using the clean latent as the supervision target yields better high-frequency recovery. The overall training procedure can be summarized as:
\begin{equation}
\begin{aligned}
\tilde{x}_{0} &= \mathrm{DU}(x_{0}), \qquad
x_{t} = \tilde{x}_{0} + \sigma_{t}\epsilon,\ \epsilon \sim \mathcal{N}(0, I), \\
\hat{x}_{0} &= x_{t} - \sigma_{t} v_{\theta}(x_{t}, t), \qquad
\mathcal{L} = \left\lVert \hat{x}_{0} - x_{0} \right\rVert^{2}.
\end{aligned}
\label{eq:hf_aware}
\end{equation}

Here, $x_{0}$ denotes the clean latent, $\mathrm{DU}(\cdot)$ is the downsample--upsample degradation operator, $\tilde{x}_{0}$ is the degraded latent, $\epsilon$ is standard Gaussian noise, $\sigma_{t}$ is the noise level at timestep $t$, $v_{\theta}(\cdot,\cdot)$ is the flow predictor, $\hat{x}_{0}$ is the reconstructed clean latent, and $\mathcal{L}$ is the training loss.

\vspace{-0.1cm}

\section{Experiments}

\subsection{Settings and Implementation Details}
We conduct experiments on Wan2.2~\cite{wan2025wanopenadvancedlargescale}, a state-of-the-art DiT-based video generation model. Wan2.2 includes a 5B variant trained only at $1280\times 704$ resolution with a VAE downsampling factor of $16\times 16\times 4$, and a 14B variant trained on a mixed-resolution dataset of $832\times 480$ and $1280\times 720$ with a VAE downsampling factor of $8\times 8\times 4$. Additional results on other DiT models are provided in the Appendix. We implement the proposed GCLFA using FlexAttention~\cite{dong2024flexattentionprogrammingmodel} in PyTorch~\cite{paszke2019pytorchimperativestylehighperformance}, leveraging efficient low-level optimizations for sparse attention. We fine-tune the parameters in the attention layers on 2.3K samples at $2752\times 1536$ resolution, generated by FLUX 1.1 Pro Ultra~\cite{labs2025flux1kontextflowmatching}, for 3K iterations in Stage~1 with the standard flow matching loss, and for another 3K iterations in Stage~2 using the loss defined in Eq.~\ref{eq:hf_aware}. Other hyper-parameters, including LoRA rank, schedulers and optimizers, follow the default settings of DiffSynth-Studio~\cite{DiffSynthStudio}. Two stage Training is conducted on a single A100 GPU and finishes in about one day. Unless otherwise specified, all inference is performed on a single A100 GPU.

The method is evaluated with VBench~\cite{huang2023vbenchcomprehensivebenchmarksuite}, which assesses both visual quality and semantic coherence. For $1920\times 1088$ (1080P) video and $3820 \times 2176$ (4K) generation, 60 prompts are randomly selected from the standard prompt suite of VBench~\cite{huang2023vbenchcomprehensivebenchmarksuite}, and the official evaluation protocol is followed. Each method generates five videos per prompt using five different random seeds. The official VBench metrics are then applied to ensure a fair comparison across all methods.

\vspace{-0.1cm}
\subsection{Main Comparisons}

We evaluate our method on Wan2.2~\cite{wan2025wanopenadvancedlargescale} across both 1080P and 4K resolutions. We consider the following baselines: (1) \textbf{Low-Level Super Resolution Methods}, including Real-ESRGAN~\cite{wang2021realesrgantrainingrealworldblind} and Upscale-A-Video~\cite{zhou2023upscaleavideotemporalconsistentdiffusionmodel}, whose base videos are derived from Wan2.2~\cite{wan2025wanopenadvancedlargescale}; (2) \textbf{Training-Free High-Resolution Generation Methods}, including I-Max~\cite{du2024imaxmaximizeresolutionpotential}, HiFlow~\cite{bu2025hiflowtrainingfreehighresolutionimage}; and (3) \textbf{Training-Based High-Resolution Generation Methods}, including CineScale~\cite{qiu2025cinescalefreelunchhighresolution} and T3-Video~\cite{zhang2025transformtrainedtransformeraccelerating}, both of which adapt models using $1080\mathrm{P}$ and $4\mathrm{K}$ real video data.

\begin{figure*}[!t]
    \centering
    \includegraphics[width=\linewidth]{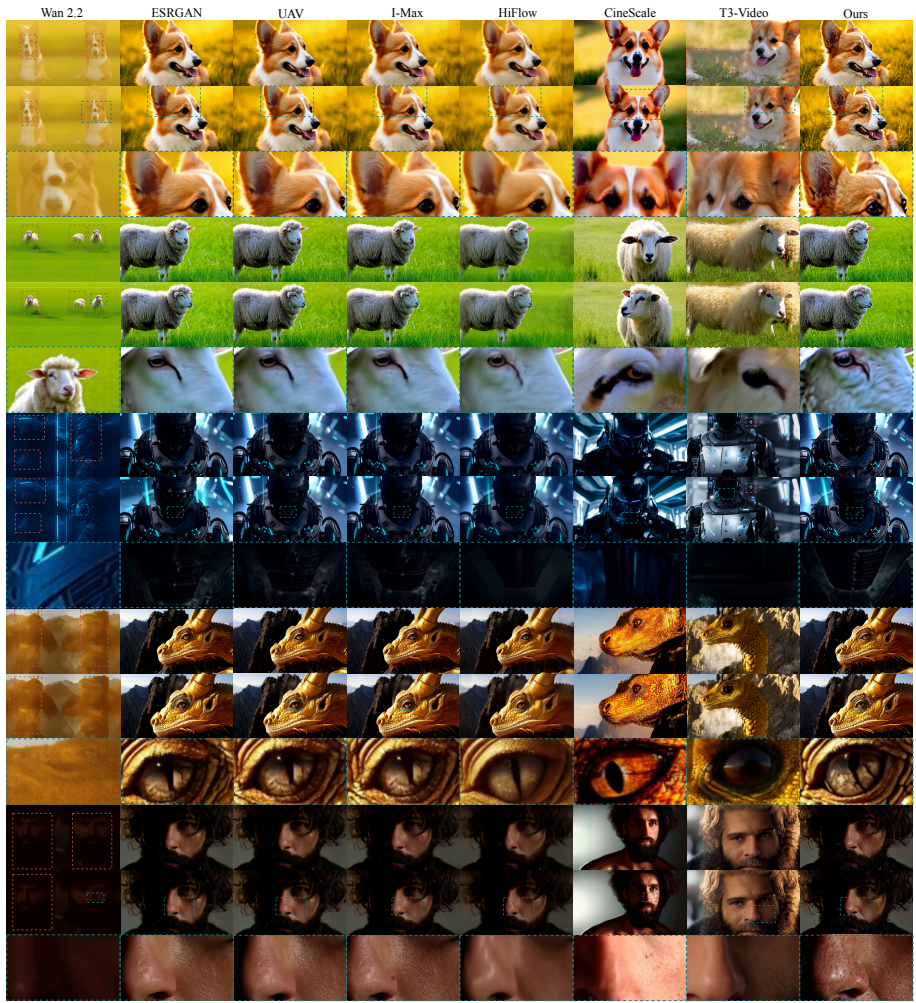}
    \vspace{-0.4cm}
    \caption{Qualitative comparison. ViBe yields high-resolution videos characterized by high-fidelity details and coherent structure. The \textcolor{red}{red boxes} highlight regions with incorrect semantics or layout. The \textcolor{myblue}{blue boxes} provide zoomed-in views.}
    \vspace{-0.4cm}
    \label{fig:qualitative}
\end{figure*}

\textbf{Quantitative comparison.}
As shown in Tabs.~\ref{tab:merged}, our method achieves competitive performance across both 1080P and 4K resolutions, particularly in terms of aesthetic quality (including layout, color richness, and harmony), imaging quality (capturing distortions such as over-exposure, noise, and blur), and overall consistency (reflecting both semantic and style alignment). These results validate the effectiveness of our method, enhancing fine-grained aesthetic details and improving the global-layout accuracy of the generated videos. Our method remains competitive to the baselines on other metrics and ranks first on the overall scores.

\textbf{Qualitative comparison.}
To visually demonstrate the superiority of our approach, as illustrated in Fig.~\ref{fig:qualitative}, we compare our method with the baselines ~\cite{wan2025wanopenadvancedlargescale,du2024imaxmaximizeresolutionpotential,zhou2023upscaleavideotemporalconsistentdiffusionmodel,wang2021realesrgantrainingrealworldblind,qiu2025cinescalefreelunchhighresolution,zhang2025transformtrainedtransformeraccelerating} designed for high-resolution visual generation on Wan2.2-5B~\cite{wan2025wanopenadvancedlargescale}. Under the same prompt, our method consistently produces the most visually appealing results, featuring the richest fine-grained details and the highest semantic consistency with the prompt. Although Real-ESRGAN~\cite{wang2021realesrgantrainingrealworldblind} and Upscale-A-Video~\cite{zhou2023upscaleavideotemporalconsistentdiffusionmodel}, two video
super-resolution methods, enhances video clarity and quality to some extent, it struggles to synthesize finer details. While I-Max~\cite{du2024imaxmaximizeresolutionpotential} and HiFlow~\cite{bu2025hiflowtrainingfreehighresolutionimage} are capable of
preserving the overall video structure in a training-free paradigm, they often synthesize low-fidelity details, leading to overly limited detail creation.
In comparison, the proposed methods consistently yields aesthetically pleasing and semantically coherent outcomes. Furthermore, our methods delivers high-resolution images of exceptional quality, outperforming leading training-based models such as CineScale~\cite{qiu2025cinescalefreelunchhighresolution} and T3-Video~\cite{zhang2025transformtrainedtransformeraccelerating}, underscoring its outstanding generative capabilities. In particular, in the fifth example, our generated video vividly presents the prompt-specific details, such as the facial details, which are barely manifested in the results of other approaches.

\begin{table}[t]
\centering
\renewcommand{\arraystretch}{1.5}
\resizebox{\textwidth}{!}{ 
\scriptsize
\begin{tabular}{l | c | c c c c c c | c}
\toprule
\toprule
\makecell[c]{\textbf{Resolution}\\\textbf{(W $\times$ H)}} & \textbf{Model} &
\makecell[c]{\textbf{Subject}\\\textbf{Consistency}} &
\makecell[c]{\textbf{Background}\\\textbf{Consistency}} &
\makecell[c]{\textbf{Motion}\\\textbf{Smoothness}} &
\makecell[c]{\textbf{Aesthetic}\\\textbf{Quality}} &
\makecell[c]{\textbf{Imaging}\\\textbf{Quality}} &
\makecell[c]{\textbf{Overall}\\\textbf{Consistency}} &
\makecell[c]{\textbf{Overall}\\\textbf{Score}} \\
\midrule
\midrule
\multirow{9}{*}{\textbf{1920 $\times$ 1088}} 
& Wan2.2~\cite{wan2025wanopenadvancedlargescale} 
& 94.9\% & 95.4\% & 98.5\% & 59.1\% & \underline{64.0\%} & 23.3\% & 72.5\% \\
& ESRGAN~\cite{wang2021realesrgantrainingrealworldblind} 
& 95.3\% & 93.8\% & 96.9\% & \underline{60.4\%} & 63.6\% & \underline{25.7\%} & \underline{72.6\%} \\
& UAV~\cite{zhou2023upscaleavideotemporalconsistentdiffusionmodel} 
& 95.2\% & 93.2\% & 96.8\% & 58.7\% & 62.5\% & 25.2\% & 71.9\% \\
& I-Max~\cite{du2024imaxmaximizeresolutionpotential} 
& \textbf{95.7\%} & 95.3\% & 97.7\% & 60.2\% & 62.0\% & 24.0\% & 72.5\% \\
& HiFlow~\cite{bu2025hiflowtrainingfreehighresolutionimage} 
& \underline{95.5\%} & 95.6\% & 97.4\% & 60.1\% & 59.6\% & 25.4\% & 72.3\% \\
& CineScale~\cite{qiu2025cinescalefreelunchhighresolution} 
& 93.9\% & \underline{96.8\%} & 97.4\% & 59.1\% & 60.7\% & 23.5\% & 71.9\% \\
& T3-Video~\cite{zhang2025transformtrainedtransformeraccelerating} 
& 94.2\% & 91.8\% & \underline{98.6\%} & 37.8\% & 35.8\% & 20.8\% & 63.2\% \\
& Ours 
& \underline{95.5\%} & \textbf{97.5\%} & \textbf{98.9\%} & \textbf{61.4\%} & \textbf{65.7\%} & \textbf{26.7\%} & \textbf{74.3\%} \\
\midrule
\midrule
\multirow{9}{*}{\textbf{3840 $\times$ 2176}} 
& Wan2.2~\cite{wan2025wanopenadvancedlargescale} 
& 94.7\% & 94.8\% & 97.1\% & 59.1\% & 33.9\% & 13.6\% & 65.5\% \\
& ESRGAN~\cite{wang2021realesrgantrainingrealworldblind} 
& \underline{95.1\%} & \textbf{97.1\%} & 97.3\% & 59.8\% & 58.3\% & 24.3\% & 72.0\% \\
& UAV~\cite{zhou2023upscaleavideotemporalconsistentdiffusionmodel} 
& 94.9\% & 96.4\% & 97.3\% & \underline{61.1\%} & 65.7\% & \underline{25.8\%} & 73.5\% \\
& I-Max~\cite{du2024imaxmaximizeresolutionpotential} 
& 94.7\% & 95.7\% & \underline{98.9\%} & 57.0\% & 61.8\% & 26.9\% & 72.5\% \\
& HiFlow~\cite{bu2025hiflowtrainingfreehighresolutionimage} 
& 95.0\% & 96.7\% & 99.1\% & 56.6\% & 54.4\% & 27.0\% & 71.5\% \\
& CineScale~\cite{qiu2025cinescalefreelunchhighresolution} 
& 94.8\% & \underline{97.0\%} & 98.2\% & 60.1\% & \textbf{66.3\%} & 25.1\% & \underline{73.6\%} \\
& T3-Video~\cite{zhang2025transformtrainedtransformeraccelerating} 
& 92.8\% & 95.4\% & 98.1\% & 60.8\% & 64.8\% & 24.7\% & 72.8\% \\
& Ours
& \textbf{95.2\%} & \textbf{97.1\%}& \textbf{99.2\%} & \textbf{61.4\%} & \underline{66.1\%} & \textbf{27.1\% }& \textbf{74.4\%} \\
\bottomrule
\end{tabular}}
\caption{Video comparison with both training-based and training-free models at 1080P and 4K resolutions. The highest value is \textbf{bold}, and the second-highest is \underline{underlined}. Overall Score is computed as the mean of the six metrics in each row.}
\label{tab:merged}
\vspace{-0.8cm}
\end{table}

\subsection{User Study}
\begin{figure*}[!t]
    \centering
    \includegraphics[width=\linewidth]{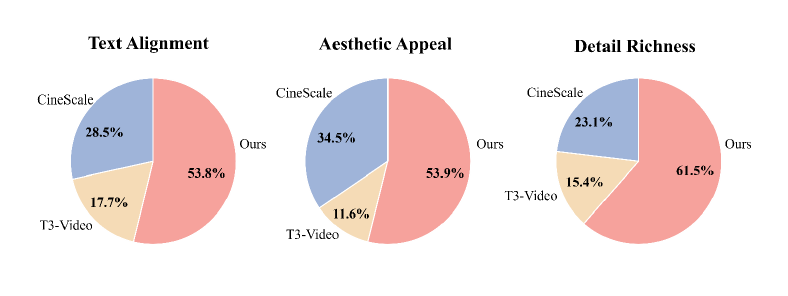}
    \vspace{-0.4cm}
    \caption{Results of user study for high-resolution video generation. Participants were expected to select the best method based on Text Alignment, Aesthetic Quality, and Video Quality.}
    \vspace{-0.1cm}
    \label{fig:pie}
\end{figure*}

\vspace{-0.1cm}
To further qualitatively evaluate the performance of our method, we conduct a human study to assess the subjective aesthetic perception of the generated videos. Specifically, we compare four methods, all based on Wan2.2-5B~\cite{wan2025wanopenadvancedlargescale}: the training-based high resolution approach CineScale~\cite{qiu2025cinescalefreelunchhighresolution} and T3-Video~\cite{zhang2025transformtrainedtransformeraccelerating} and our proposed method. All methods are evaluated using the same prompts from VBench~\cite{huang2023vbenchcomprehensivebenchmarksuite}. During the study, each participant was presented with the generated videos in a randomized order and expected to select the best video based on three criteria: Aesthetic Appeal, Detail Richness, and Text Alignment. Aesthetic Appeal referred to the overall visual attractiveness and artistic quality of the video, while Detail Richness evaluated the level of fine-grained details present in the video frames. Text Alignment assessed how well the generated content adhered to the textual input, ensuring that the video accurately reflected the provided prompts. In total, 35 participants took part in the study and provided their subjective preferences. As shown in Fig.~\ref{fig:pie}, our method consistently receives the highest preference across all the metrics.

\subsection{Ablation Study}\label{sec:ablation}
We conduct controlled ablations to quantify the contribution of each component in our framework, including (i) Relay LoRA, (ii) Global-Coarse-Local-Fine-Attention (GCLFA), and (iii) High-Frequency-Awareness-Training-Objective (HFATO). Quantitative results are reported in Tab.~\ref{tab:ablations}, and qualitative comparisons are shown in Fig.~\ref{fig:ablation}. we draw the following conclusions regarding
the contribution of each component:

\begin{figure*}[!t]
    \centering
    \includegraphics[width=\linewidth]{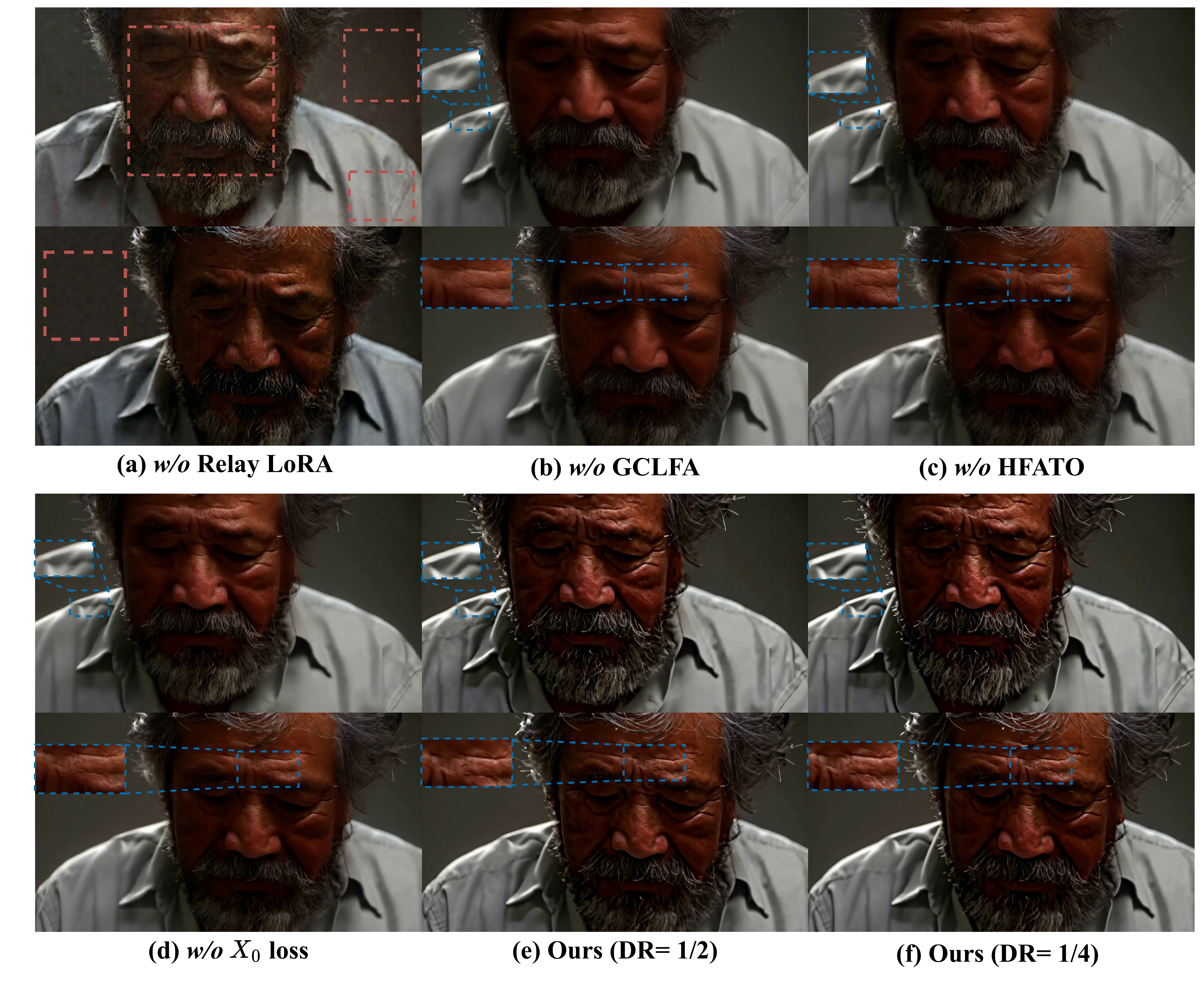}
    \vspace{-0.4cm}
    \caption{Qualitative ablations. We perform controlled comparisons of our method with alternative variants under controlled settings. Best viewed zoomed in. The \textcolor{red}{red boxes} highlight regions with more noticeable noise, and the \textcolor{myblue}{blue boxes} indicate the corresponding zoomed-in views.}
    \vspace{-0.4cm}
    \label{fig:ablation}
\end{figure*}

\textbf{Relay LoRA.}
Relay LoRA is crucial for suppressing the noise induced by naïve high-resolution image fine-tuning. Compared to direct high-resolution LoRA adaptation, Relay LoRA yields markedly cleaner and more temporally stable videos (Fig.~\ref{fig:ablation}(a)$\rightarrow$(b)). Without Relay LoRA, the model either fails to maintain coherent content at the target resolution or suffers from severe instability, establishing Relay LoRA as the foundation for reliable ultra-high-resolution generation.

\textbf{Global-Coarse-Local-Fine-Attention (GCLFA).}
GCLFA further improves fine-grained details while preserving global structure (Fig.~\ref{fig:ablation}(b)$\rightarrow$(c)). Removing GCLFA consistently degrades visual richness and semantic integrity, indicating that coarse global context combined with local fine attention is essential for high-resolution detail synthesis.

\textbf{High-Frequency-Awareness-Training-Objective (HFATO).}
HFATO consists of two decoupled factors: latent degradation and $x_0$-reconstruction supervision. When we apply only the downsample--upsample degradation without the $x_0$ loss, the model gains limited detail enhancement (Fig.~\ref{fig:ablation}(c)$\rightarrow$(d)). Adding the $x_0$ reconstruction loss substantially boosts high-frequency details and improves robustness (Fig.~\ref{fig:ablation}(d)$\rightarrow$(e)), suggesting that explicitly supervising the recovered clean latent is key to stable detail amplification.

\begin{table*}[t]
\centering
\renewcommand{\arraystretch}{1.28}
\resizebox{0.97\textwidth}{!}{ 
\scriptsize  
\begin{tabular}{l | c c c c c c | c}
\toprule
\toprule
\textbf{Methods} &
\makecell[c]{\textbf{Subject}\\\textbf{Consistency}} &
\makecell[c]{\textbf{Background}\\\textbf{Consistency}} &
\makecell[c]{\textbf{Motion}\\\textbf{Smoothness}} &
\makecell[c]{\textbf{Aesthetic}\\\textbf{Quality}} &
\makecell[c]{\textbf{Imaging}\\\textbf{Quality}} &
\makecell[c]{\textbf{Overall}\\\textbf{Consistency}} &
\makecell[c]{\textbf{Overall}\\\textbf{Score}} \\
\midrule
\midrule

\textit{w/o} Relay LoRA       & 94.6\% & \textbf{97.9\%} & 98.3\% & 46.7\% & 37.4\% & 21.1\% & 66.0\% \\
\textit{w/o} GCLFA                                   & 94.9\% & 95.4\% & 99.0\% & 52.2\% & 49.0\% & 24.3\% & 69.1\% \\
\textit{w/o} HFATO                               & 95.0\% & 94.2\% & 99.1\% & 53.7\% & 59.0\% & 24.0\% & 70.8\% \\
\textit{w/o} $X_0$ Loss                                  & \textbf{96.3\%} & 95.3\% & \underline{99.2\%} & 57.3\% & 59.2\% & \underline{26.7\%} & 72.3\% \\
Ours (DR = 1/2)                              & \underline{95.2\%} &  97.1\% & \underline{99.2\%} & \textbf{61.4\%} & \textbf{66.1\%} & \textbf{27.1\%} & \textbf{74.4\%} \\
Ours (DR = 1/4)                               & 95.0\% & \underline{97.2\%} & \textbf{99.3\%}  & \underline{61.3\%} & \underline{65.6\%} & 26.5\% & \underline{74.2\%} \\
\bottomrule
\bottomrule
\end{tabular}
}
\caption{Quantitative ablations. Ours Methods achieves the optimal overall video quality. The highest value is \textbf{bold}, and the second-highest is \underline{underlined}.}
\label{tab:ablations}
 \vspace{-0.5cm}

\end{table*}

\textbf{Effect of the Downsample Ratio (DR).}
In GCLFA, we investigate two downsample ratios: $1/2$ and $1/4$. As shown in Tab.~\ref{tab:ablations}, both ratios are competitive in terms of aesthetic and imaging quality. Additionally, both ratios generate coherent content (Fig.~\ref{fig:ablation}(e)$\rightarrow$(f)), with the $1/4$ downsample ratio further enhancing fine details, such as wrinkles on the elderly's skin, especially when zoomed in. Consequently, we recommend a downsample ratio of $1/2$ for more robust results, and $1/4$ for results with greater detail.

\subsection{Empirical Study}
\begin{figure*}[!t]
    \centering
    \includegraphics[width=\linewidth]{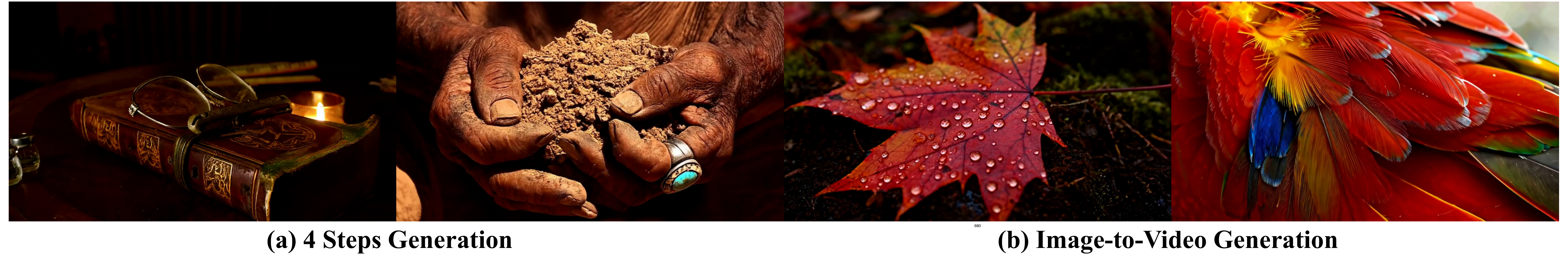}
    \vspace{-0.5cm}
    \caption{\textbf{Left}: Illustration 4 steps generation results integrated with our framework. \textbf{Right}: Illustration Image-to-Video generation results integrated with our framework.}
    \vspace{-0.5cm}
    \label{fig:generalization}
\end{figure*}

To further validate the generalization of our framework, we conduct an empirical study across different scenarios. 
As illustrated in Fig.~\ref{fig:generalization}(a), experiments show that our $\mathrm{LoRA}_2$ can be directly integrated into few steps distilled models~\cite{li2025magicmotion} to significantly accelerate inference speed while maintaining high-fidelity visual details with negligible degradation. 
Furthermore, as illustrated in Fig.~\ref{fig:generalization}(b), experiments show that our $\mathrm{LoRA}_2$ can be directly integrated into Image-to-Video (I2V) generation tasks to upgrading existing I2V models to synthesize ultra-high-resolution content.

\section{Conclusion}

In this paper, we introduce ViBe, a novel image-based training paradigm designed for ultra-high-resolution video generation. By introducing \textit{Relay LoRA}, ViBe effectively addresses challenges in naively fine-tuning pre-trained text-to-video diffusion transformers, which are originally trained only on low-resolution scales. Our method mitigates issues such as residual noise when using high-resolution images. Additionally, ViBe incorporates \textit{Global-Coarse-Local-Fine Attention}, which enhances visual details while preserving semantic consistency. Furthermore, we propose the \textit{High-Frequency-Awareness-Training-Objective} to improve the model's capability to handle high-frequency details. Experimental results demonstrate that ViBe outperforms existing methods in ultra-high-resolution video generation, achieving superior video quality.

\par\vfill\par
\clearpage  


%
%
\bibliographystyle{splncs04}
\bibliography{main}

\newpage

\section{Supplementary Material}

\begin{figure*}[h]
    \vspace{-1cm}
    \centering
    \includegraphics[width=0.92\linewidth]{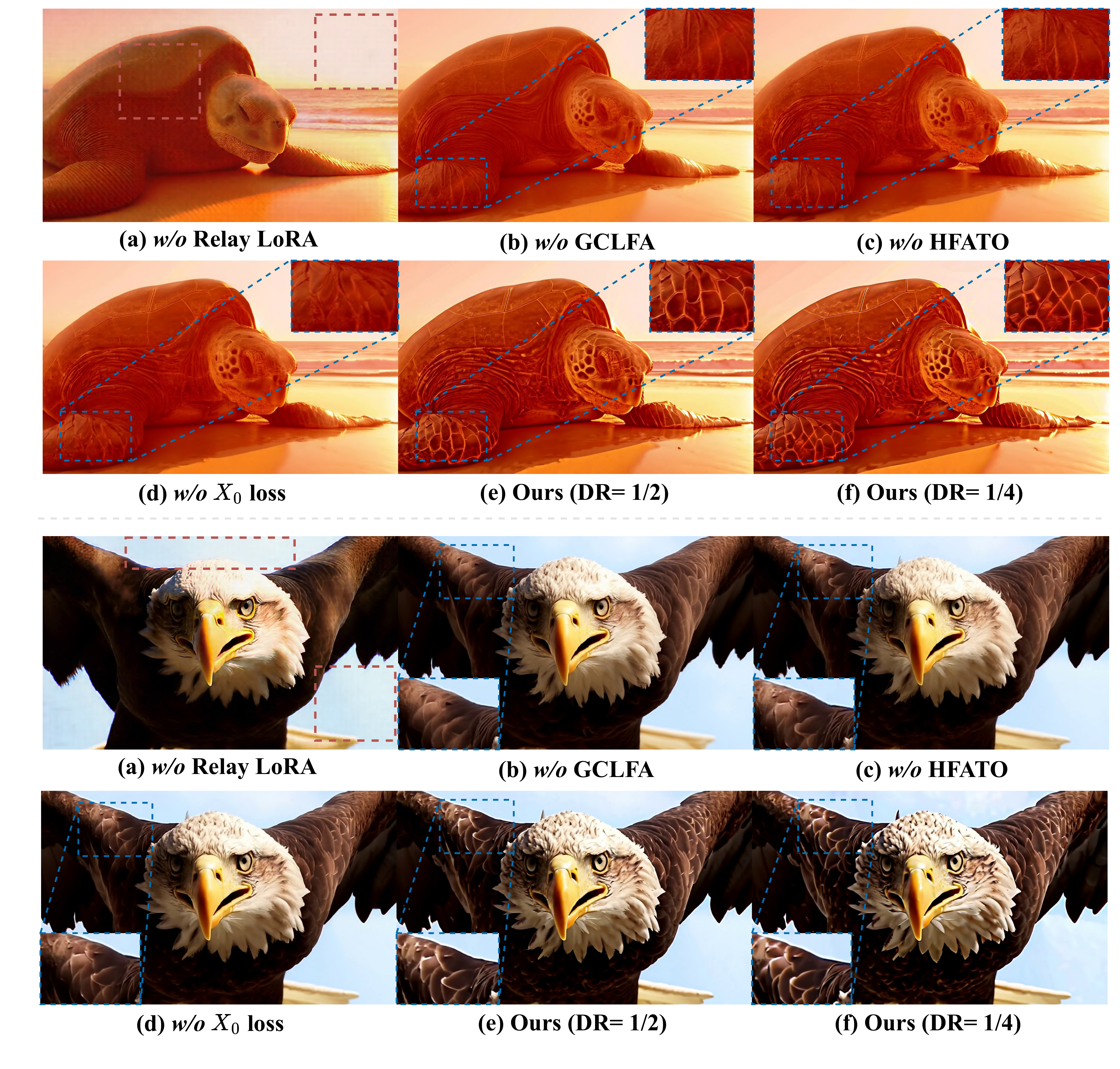}
    \vspace{-0.5cm}
    \caption{More Qualitative ablations. We perform controlled comparisons of our method with alternative variants under controlled settings. Best viewed zoomed in. The \textcolor{red}{red boxes} highlight regions with more noticeable noise, and the \textcolor{myblue}{blue boxes} indicate the corresponding zoomed-in views.}
    \vspace{-0.5cm}
    \label{fig:more ablation}
\end{figure*}

\vspace{-0.5cm}
\textbf{Ablation Study.} 
As illustrated in Fig.~\ref{fig:more ablation}, we extend the motivational examples from Fig.~\ref{fig:ablation} with additional cases to conduct a comprehensive ablation study, further validating the effectiveness of our proposed components. 
Comparing Fig.~\ref{fig:more ablation}(a) and (b), we observe that Fig.~\ref{fig:more ablation}(a) suffers from noise artifacts stemming from the modality gap between pre-training video data and fine-tuning image data. 
In contrast, our proposed Relay LoRA (Fig.~\ref{fig:more ablation}(b)) effectively mitigates this gap in a decoupled manner while enabling spatial extrapolation, thereby demonstrating its efficacy. 
Furthermore, the comparison between Fig.~\ref{fig:more ablation}(b) and (c) reveals that the GCLFA module further enhances the generation of fine visual details. 
We then investigate the two decoupled factors within HFATO: latent degradation and $x_0$-reconstruction supervision. 
As shown in Fig.~\ref{fig:more ablation}(c)$\rightarrow$(d), applying latent degradation without the $x_0$ loss yields only marginal detail enhancement. 
However, incorporating $x_0$ reconstruction significantly boosts high-frequency details (Fig.~\ref{fig:more ablation}(d)$\rightarrow$(e)), suggesting that explicit supervision of the recovered clean latent is essential for stable detail amplification.

\textbf{Effect of Pooling in Global-Coarse-Local-Fine Attention (GCLFA).} 
To validate the necessity of the global coarse branch in GCLFA, we replace it with local attention in Stage 2 and fine-tune the model using the same Relay LoRA and HFATO for fairness. As shown in Fig.~X(a), this modification tends to produce repetitive patterns in high-resolution video generation. In contrast, our proposed GCLFA preserves fine visual details while enhancing the robustness of the global structure, demonstrating the effectiveness of the global coarse component.

\begin{figure*}[h]
    \vspace{-0.5cm}
    \centering
    \includegraphics[width=1\linewidth]{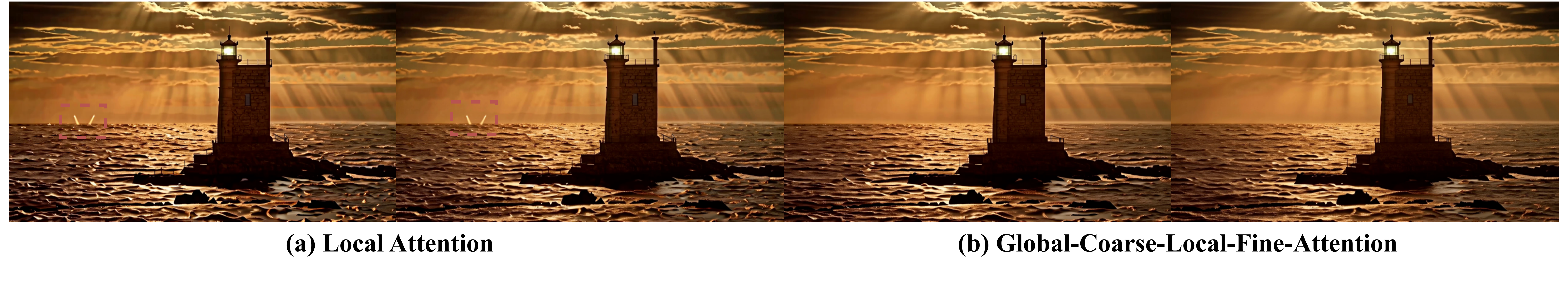}
    \vspace{-0.2cm}
    \caption{A fair comparison of Local Attention and GCLFA. We perform controlled comparisons of our method with attnetion variants under controlled settings. Best viewed zoomed in. The \textcolor{red}{red boxes} highlight regions with incorrect semantics or layout.}
    \vspace{-0.5cm}
    \label{fig:attn}
\end{figure*}

\textbf{Evaluation by VLM.}
We also provide quantitative comparison of the widely adopted GPT-based evaluation, are summarized in Tab.~\ref{tab:model performance}. To ensure an objective and fine-grained assessment, we designed a comprehensive evaluation protocol. The following prompt was provided to the VLM (e.g., GPT-4o) to evaluate the outputs of eight different models based on a unified generation task:

\begin{quote}
\textit{``Please evaluate the results generated by 8 different models. All models used the same prompt: `A lovable Welsh Corgi sitting on sunlit grass, framed in a tight close-up on its face and upper body, filling most of the frame. Emphasize thick, fluffy fur with clear individual strands, soft undercoat volume, subtle color variation in the tan-and-white coat, and tiny flyaway hairs catching the light. Big round eyes with deep, glossy reflections and crisp catchlights; detailed nose texture with gentle moisture shine; visible whiskers and fine muzzle fur. The corgi’s ears perk up slightly and its tongue peeks out in a cheerful expression. Warm golden-hour sunlight creates soft rim light along the fur edges; shallow depth of field with creamy bokeh background. Ultra-detailed textures, high realism, cinematic look, 8K resolution.' Please provide scores (1-10) across three dimensions: \textbf{Aesthetic Appeal}, \textbf{Detail Richness}, and \textbf{Text Alignment}.''}
\end{quote}

\begin{tcolorbox}[
    title=Evaluation by VLM,
    colframe=black!20,
    colback=white,
    coltitle=black,
    fonttitle=\bfseries,
    breakable
]
\begin{table}[H]
\vspace{-0.95cm}
\centering
\setlength{\tabcolsep}{1.0pt}
\resizebox{\linewidth}{!}{
\renewcommand{\arraystretch}{1.08}
\begin{tabular}{l|cccc} 
\toprule
\textbf{Model} & 
\makecell{Aesthetic\\Appeal} & 
\makecell{Detail\\Richness} & 
\makecell{Text\\Alignment} &
\makecell{Overall\\Scores} \\
\midrule
\textbf{Wan2.2~\cite{wan2025wanopenadvancedlargescale}} & 5.25 & 4.50 & 5.25 & 5.00 \\
\textbf{Real-ESRGAN~\cite{wang2021realesrgantrainingrealworldblind}} & 7.75 & 7.00 & 8.50 & 7.75 \\
\textbf{UAV~\cite{zhou2023upscaleavideotemporalconsistentdiffusionmodel}} & 7.50 & 7.25 & 8.50 & 7.75 \\
\textbf{I-Max~\cite{du2024imaxmaximizeresolutionpotential}} & 8.25 & 7.75 & 9.25 & 8.42 \\
\textbf{HiFLow~\cite{bu2025hiflowtrainingfreehighresolutionimage}} & 8.00 & 8.00 & 9.25 & 8.42 \\
\textbf{CineScale~\cite{qiu2025cinescalefreelunchhighresolution}} & 7.50 & 6.75 & 7.50 & 7.25 \\
\textbf{T3-Video~\cite{su2023roformerenhancedtransformerrotary}} & 6.75 & 6.75 & 7.25 & 6.92 \\
\rowcolor[gray]{0.9} \textbf{Ours} & 8.00 & 8.25 & 9.25 & 8.50 \\
\bottomrule
\end{tabular}
}
\caption{Model Performance Averages for Aesthetic Quality, Detail Richness, and Text Alignment evaluated.}
\label{tab:model performance}
\vspace{-0.6cm}
\end{table}

\end{tcolorbox}

\begin{tcolorbox}[
    title=Qualitative Feedback from VLM,
    colframe=black!20,
    colback=white,
    coltitle=black,
    fonttitle=\bfseries,
    breakable
]
\begin{table}[H]
\vspace{-0.95cm}
\centering
\small
\renewcommand{\arraystretch}{1.2}
\resizebox{\linewidth}{!}{
\begin{tabular}{l|p{10.5cm}}
\toprule
\textbf{Model ID} & \textbf{Qualitative Assessment and Response} \\
\midrule
\textbf{Model 1} & The image suffers from a heavy yellowish tint and significant motion-like blur, lacking the necessary contrast for high-fidelity output. \\
\hline
\textbf{Model 2} & Features standard composition with soft lighting; however, the wool texture appears overly smooth and lacks individual fiber definition. \\
\hline
\textbf{Model 3} & Exhibits performance nearly identical to Model 2, with only minor variations in color temperature. \\
\hline
\textbf{Model 4} & Demonstrates excellent rendering with enhanced depth in lighting layers; fine wool textures begin to manifest clearly. \\
\hline
\textbf{Model 5} & Closely follows Model 4's quality, particularly excelling in the execution of the golden rim lighting effect as specified in the prompt. \\
\hline
\textbf{Model 6} & Adopts a frontal perspective, but the lighting is somewhat stiff and fails to capture the requested "cinematic" atmosphere. \\
\hline
\textbf{Model 7} & The side-lying posture and framing deviate from the specific "standing" and "tight close-up" requirements of the text prompt. \\
\hline
\rowcolor[gray]{0.9} 
\textbf{Model 8} & \textbf{The superior performer. It achieves full adherence to all descriptive constraints, demonstrating exceptional textural fidelity and professional-grade realism.} \\
\bottomrule
\end{tabular}
}
\caption{Detailed qualitative breakdown of VLM-generated feedback for the representative sheep case.}
\label{tab:qualitative_feedback}
\vspace{-0.6cm}
\end{table}
\end{tcolorbox}

\begin{figure*}[h]
    \centering
    \includegraphics[width=0.95\linewidth]{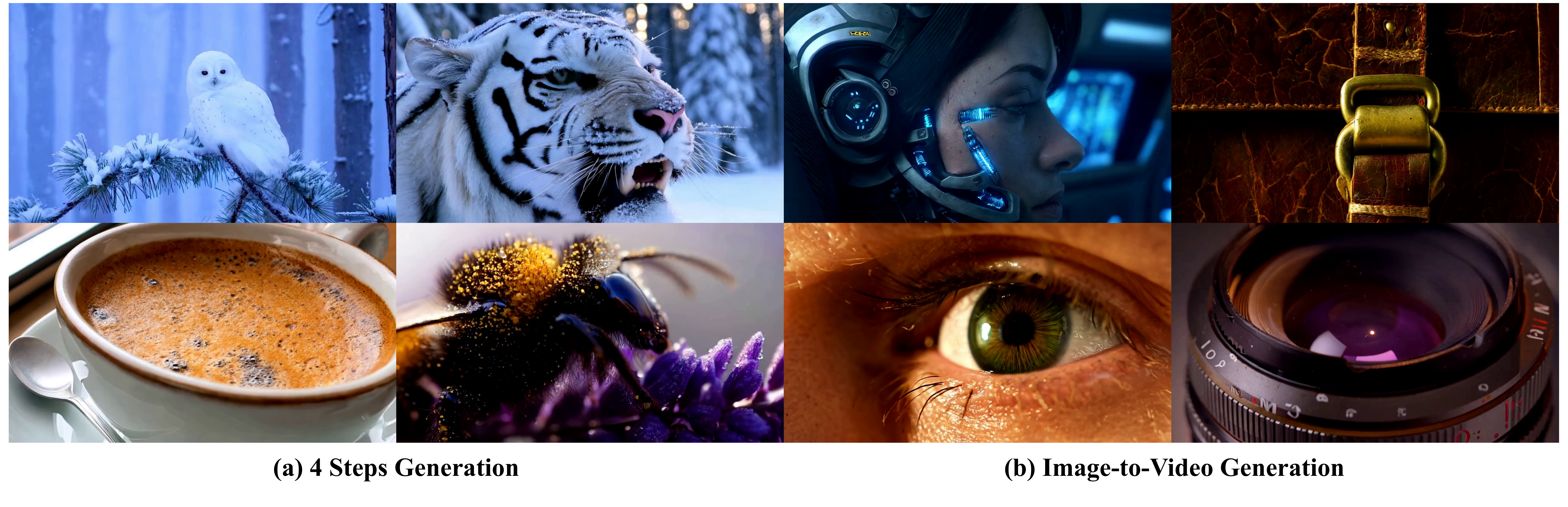}
    \vspace{-0.5cm}
    \caption{\textbf{Left}: Illustration 4 steps generation results integrated with our framework. \textbf{Right}: Illustration Image-to-Video generation results integrated with our framework.}
    \vspace{-0.4cm}
    \label{fig:i2v}
\end{figure*}

\newpage
\textbf{The Generalization of Our Method.}
As shown in Fig.~\ref{fig:i2v}(a), ViBe can be readily integrated into few-step distilled models~\cite{li2025magicmotion}, enabling efficient and high-fidelity video generation. Moreover, as illustrated in Fig.~\ref{fig:i2v}(b), ViBe is also compatible with image-to-video (I2V) generation, allowing existing I2V models to be upgraded for synthesizing ultra-high-resolution content. 

\begin{figure*}[h]
    \centering
    \includegraphics[width=0.95\linewidth]{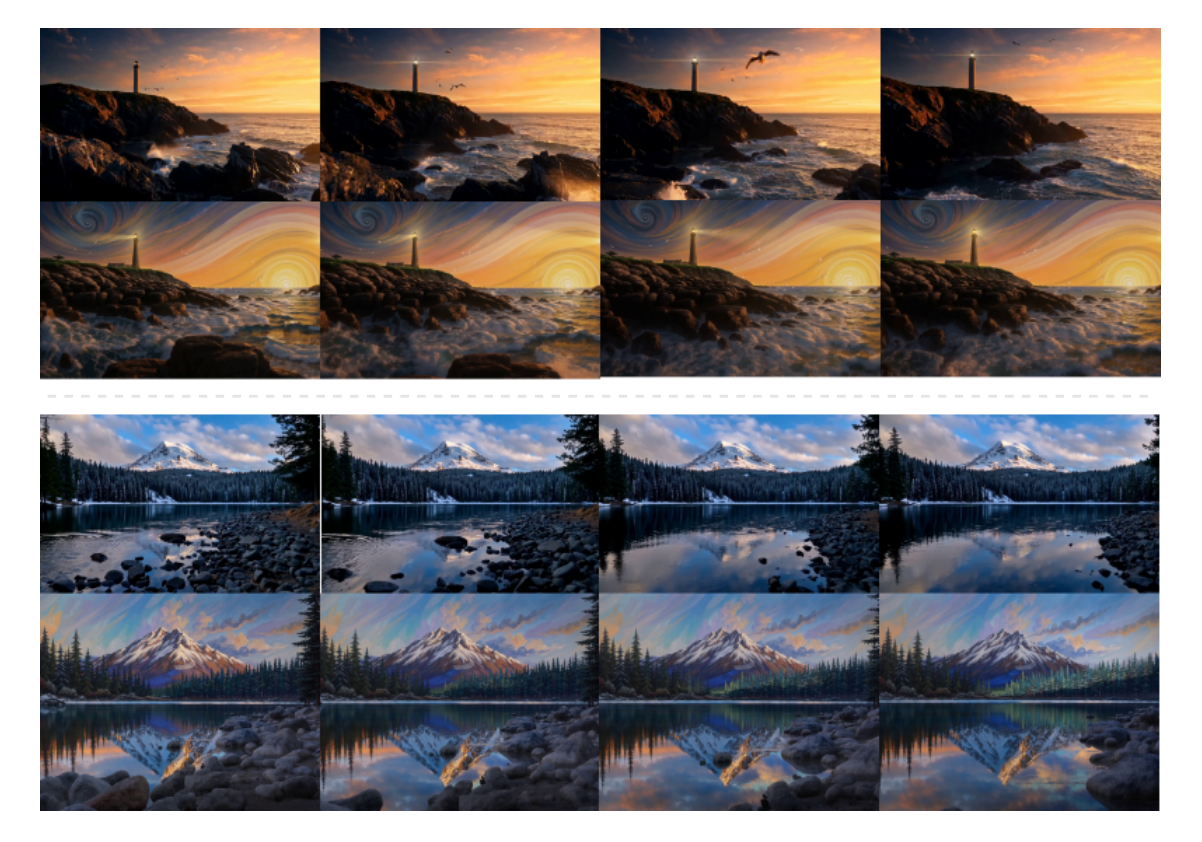}
    \vspace{-0.5cm}
    \caption{Two examples of style transfer with ViBe. In each case, the top row shows the original video, while the bottom row presents the stylized result.}
    \label{fig:tranfer}
\end{figure*}

In addition, under the coarse-to-fine pipeline, ViBe naturally supports style transfer during the video-to-video refinement stage by simply modifying the prompt, \textit{e.g.}, transforming a landscape scene into a Van Gogh style painting, while still maintaining strong visual quality, as shown in Fig.~\ref{fig:tranfer}. 

\begin{figure*}[!h]
    \vspace{-0.5cm}
    \centering
    \includegraphics[width=1\linewidth]{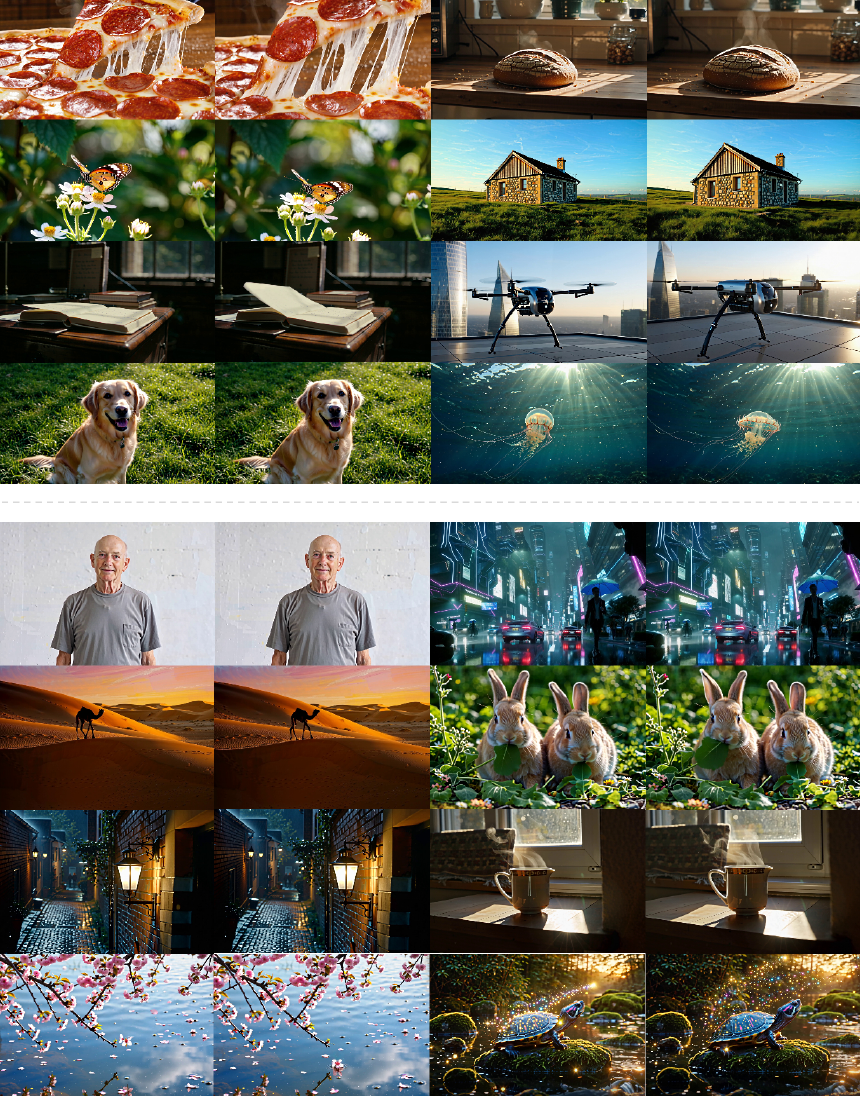}
    \vspace{-0.5cm}
    \caption{Qualitative results of 1080P and 2K generated by our method ViBe integrated with Wan2.2~\cite{wan2025wanopenadvancedlargescale}.}
    \label{fig:case1}
\end{figure*}

\begin{figure*}[!h]
    \centering
    \includegraphics[width=1\linewidth]{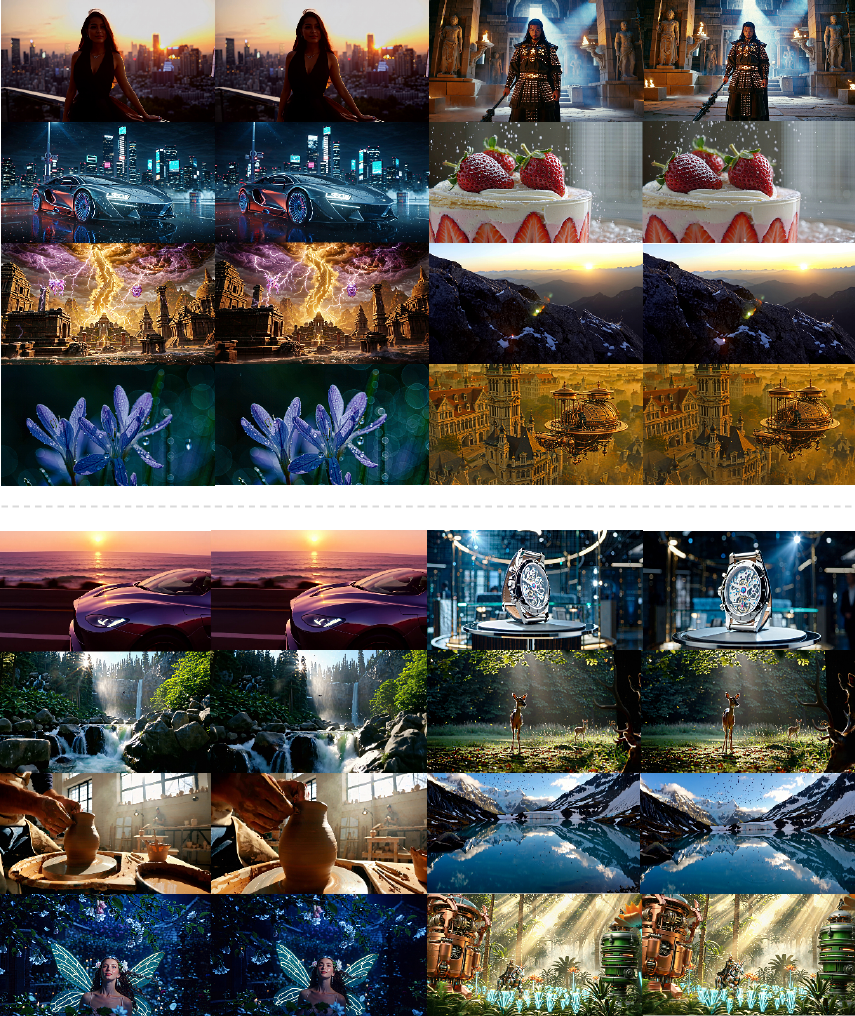}
    \caption{Qualitative results of 3K and 4K generated by our method ViBe integrated with Wan2.2~\cite{wan2025wanopenadvancedlargescale}.}
    \label{fig:case2}
\end{figure*}

\textbf{Case Study.} 
As shown in Fig.~\ref{fig:case1} and~\ref{fig:case2}, we adapt ViBe to Wan2.2~\cite{wan2025wanopenadvancedlargescale} to generate different resolution videos, i.e. 1080P, 2K, 3K, 4K. These qualitative results demonstrate that our method can consistently preserve coherent global semantic structure while producing fine-grained details across diverse scenarios, including landscapes, objects, and both static and dynamic scenes.

\vfill
\newpage
\vfill
\vfill

\textbf{VRAM Analysis.} We report the VRAM consumption (in GB) for both training and inference stages under different resolutions and downsampling ratios.

\textit{Inference:}  
\textbf{Pool 2:} 1080P: 34.54, 2K: 38.74, 3K: 64.71, 4K: 74.82  
\textbf{Pool 4:} 1080P: 34.54, 2K: 38.74, 3K: 64.69, 4K: 74.82  

\textit{Training:}  
\textbf{Stage 1:} 27.99  
\textbf{Stage 2:} Pool 2: 46.28, Pool 4: 46.28

\textbf{Implementation Details.} 
We provide the comprehensive hyperparameter configurations used in our training and inference pipeline.

\begin{tcolorbox}[
    title=Hyperparameters for Stage 1 Fine-tuning,
    colframe=black!15,
    colback=white,
    coltitle=black,
    fonttitle=\small\bfseries,
    left=5pt, right=5pt, top=5pt, bottom=5pt,
    breakable
]
\small
\begin{itemize}[leftmargin=*, itemsep=2pt]
    \item \texttt{Resolution (H $\times$ W)} = $704 \times 1280$
    \item \texttt{Number of Frames} = 1
    \item \texttt{Learning Rate} = $1 \times 10^{-4}$
    \item \texttt{Number of Epochs} = 2
    \item \texttt{LoRA Rank} = 32
    \item \texttt{LoRA Target Modules} = q, k, v, o, ffn.0, ffn.2
    \item \texttt{LoRA Base Model} = DiT
    \item \texttt{Batch Size} = 1
    \item \texttt{Dataset Repeat} = 1
    \item \texttt{Save Steps} = 500
\end{itemize}
\end{tcolorbox}

\begin{tcolorbox}[
    title=Hyperparameters for Stage 2 Fine-tuning,
    colframe=black!15,
    colback=white,
    coltitle=black,
    fonttitle=\small\bfseries,
    left=5pt, right=5pt, top=5pt, bottom=5pt,
    breakable
]
\small
\begin{itemize}[leftmargin=*, itemsep=2pt]
    \item \texttt{Resolution (H $\times$ W)} = $1536 \times 2752$
    \item \texttt{Number of Frames} = 1
    \item \texttt{Learning Rate} = $1 \times 10^{-4}$
    \item \texttt{Number of Epochs} = 2
    \item \texttt{LoRA Rank} = 32
    \item \texttt{LoRA Target Modules} = q, k, v, o, ffn.0, ffn.2
    \item \texttt{LoRA Base Model} = DiT
    \item \texttt{Batch Size} = 1 
    \item \texttt{Dataset Repeat} = 1
    \item \texttt{Save Steps} = 500
\end{itemize}
\end{tcolorbox}

\begin{tcolorbox}[
    title=Hyperparameters for Inference,
    colframe=black!15,
    colback=white,
    coltitle=black,
    fonttitle=\small\bfseries,
    left=5pt, right=5pt, top=5pt, bottom=5pt,
    breakable
]
\small
\begin{itemize}[leftmargin=*, itemsep=2pt]
    \item \texttt{num inference steps} = 50
    \item \texttt{denoising strength} = 0.7
    \item \texttt{frames per second} (FPS) = 15
    \item \texttt{quality} = 5
    \item \texttt{guidance scale} = 5.0
    \item \texttt{max sequence length} = 512
\end{itemize}
\end{tcolorbox}

\newpage
\begin{tcolorbox}[title=Prompts for Video Generation,
                  colframe=black!20,
                  colback=white,
                  coltitle=black,
                  fonttitle=\bfseries,
                  breakable]
\footnotesize
\textbf{Figure 1 – Video grid.}\\
Images are arranged starting from the top-left corner, ordered row-wise from left to right and from top to bottom. 
The prompts are listed below in the same order. 
\begin{enumerate}[leftmargin=1.2em,itemsep=2pt]
  \item A cinematic ultra-detailed food photography shot of a Japanese ramen bowl, rich broth with visible oil droplets, perfectly cooked noodles, sliced chashu pork with marbling details, soft-boiled egg with glossy yolk, steam rising naturally, dramatic side lighting, shallow depth of field, macro food photography, realistic textures, dark moody background, professional food styling, photorealistic, 8K resolution, masterpiece quality.

  \item A majestic silver-gray wolf standing on a snowy mountain ridge, thick and fluffy fur blowing in the cold wind, individual hair strands sharply visible, piercing amber eyes staring forward, breath visible in the freezing air, dramatic cloudy sky with soft sunlight breaking through, cinematic rim lighting highlighting the fur texture, ultra-realistic animal anatomy, shallow depth of field, sharp focus on the wolf’s face, high contrast, 8k ultra-high resolution, epic wildlife photography style, masterpiece quality.

  \item A detailed scene of an astronaut standing in a dense jungle, the astronaut's white space suit contrasting with the vibrant green foliage around. The jungle is alive with towering trees covered in thick vines, and exotic plants with bright colors peek through the shadows. Soft beams of light filter through the dense canopy above, casting intricate shadows on the astronaut’s suit and the surrounding flora. The astronaut’s visor reflects the jungle environment, adding depth to the scene. The overall color palette is cold and muted, with shades of green, blue, and gray dominating the image, creating a surreal contrast between the futuristic suit and the wild, untamed jungle. High-resolution 8k quality, highly detailed textures, and atmospheric lighting enhance the realism and intrigue of the scene.

  \item A futuristic armored warrior standing in the center of a high-tech megacity, wearing advanced exoskeleton armor made of dark metallic alloy and glowing energy lines, intricate mechanical joints and layered armor plates, subtle scratches and wear on the surface, helmet with illuminated visor and holographic interface, powerful and confident stance, cyberpunk cityscape in the background with towering skyscrapers, neon lights, holographic billboards, and flying vehicles, dynamic light reflections on the armor, cinematic dramatic lighting with strong rim light and volumetric light beams, slight mist and particles in the air, ultra-realistic textures, sharp focus on the armored character, shallow depth of field, epic cinematic composition, hyper-detailed, 8k ultra-high resolution, masterpiece quality, sci-fi concept art style.

  \item Surreal portrait of a young woman whose face merges with a complex arrangement of vivid flowers, bug’s-eye macro viewpoint from within petals emphasizing scale and perspective, vines and blossoms weaving seamlessly into translucent skin with freckles and delicate textures, diffused lighting casting soft shadows across blooms and contours, shallow depth of field with silky bokeh and fantastical elegance.

  \item A vast mountainous landscape at sunrise, towering cliffs and layered peaks stretching into the distance, golden sunlight breaking through thick clouds and illuminating mist-filled valleys.Ancient stone paths winding along the mountainside, sparse pine trees swaying gently in the wind.Cinematic wide-angle shot, volumetric lighting, soft fog, epic scale, ultra-detailed terrain textures, natural color grading, majestic and tranquil atmosphere, high realism, movie scene quality.
  
  \item Under the deep ocean, a ghostly ship floats between glowing jellyfish and spiral corals, its hull illuminated by glowing marine life. The blue fire from distant underwater volcanoes lights up the water, with every detail—from reflections to tiny glowing organisms—coming alive in 8K resolution, creating a dreamlike atmosphere with a haunting glow.

  \item A fluffy red fox sitting in an autumn forest, extremely thick and soft fur with fine individual hairs, vivid orange and white color transitions, alert eyes reflecting ambient light, fallen leaves covering the ground, soft morning sunlight filtering through trees, cinematic shallow depth of field, ultra-sharp fur texture, realistic animal proportions, 8k ultra-high resolution, hyper-realistic wildlife photography, cozy yet detailed atmosphere.

\end{enumerate}

\vspace{6pt}
\textbf{Figure 3 – Video Examples.}
\begin{enumerate}[leftmargin=1.2em,itemsep=2pt]
  \item A kind elderly man in a light-colored shirt with short silvery-white hair, captured in the natural motion of picking up a cup of hot tea from a wooden table and beginning to stand up slightly. The movement is fluid and coherent, with subtle folds in his sleeves shifting naturally with the action. He has a gentle smile on his face. Set in a tidy living room with a bookshelf in the background, bathed in warm, natural indoor light. Real photography style, shallow depth of field, subtle motion blur for a candid snapshot feel, cinematic lighting, high quality, and crisp details.
\end{enumerate}

\vspace{6pt}
\textbf{Figure 5 – Video Examples.}
\begin{enumerate}[leftmargin=1.2em,itemsep=2pt]
  \item A lovable Welsh Corgi sitting on sunlit grass, framed in a tight close-up on its face and upper body, filling most of the frame. Emphasize thick, fluffy fur with clear individual strands, soft undercoat volume, subtle color variation in the tan-and-white coat, and tiny flyaway hairs catching the light. Big round eyes with deep, glossy reflections and crisp catchlights; detailed nose texture with gentle moisture shine; visible whiskers and fine muzzle fur. The corgi’s ears perk up slightly and its tongue peeks out in a cheerful expression. Warm golden-hour sunlight creates soft rim light along the fur edges; shallow depth of field with creamy bokeh background. Ultra-detailed textures, high realism, cinematic look, 8K resolution.
  
  \item A cute fluffy sheep standing on a fresh green meadow, looking around curiously from side to side. Medium-close to close-up framing on the sheep’s head and upper body, filling most of the frame. Emphasize abundant wool: dense, springy curls with individual fibers visible, subtle color variation from creamy white to warm beige, tiny bits of grass caught in the fleece, and soft rim light outlining the wool volume. The sheep’s eyes are bright and expressive with crisp catchlights; detailed nose and mouth texture; small ear flicks and gentle head turns add natural motion. Sunlit grass blades in the foreground show fine texture and dew sparkle; background meadow fades into creamy bokeh. Natural cinematic lighting, ultra-detailed textures, high realism, 8K resolution.
  
  \item A futuristic humanoid combat robot standing in a neon-lit sci-fi corridor. Extreme close-up / tight close-up on the robot’s helmet, chest plate, and shoulder armor, filling most of the frame. Emphasize heavy mech plating, layered armor panels, exposed mechanical joints, micro-scratches, brushed metal grain, carbon-fiber inserts, and tiny engraved serial markings. Glowing cyan energy lines pulse beneath translucent armor seams; small vents release faint heat haze; miniature pistons and cables subtly move with idle motion. Crisp specular highlights and detailed reflections of holographic UI elements across the polished surfaces. Volumetric light beams, floating dust particles, cinematic shallow depth of field with background blurred into soft bokeh. Ultra-detailed textures, high-tech atmosphere, hyper-real close-up, 8K resolution.

  \item An awe-inspiring golden dragon perched on a jagged rocky cliff high in the mountains, its molten-gold scales catching the sunlight. Extreme close-up / tight close-up on the dragon’s head and upper neck, filling most of the frame. Emphasize razor-sharp eyes with reflective highlights, subtle iris patterns, detailed eyelids and brow ridges, fine facial scale micro-patterns, tiny nicks and weathered scratches on the scales, and faint dust/pollen particles clinging to the surface. Long whiskers and horn ridges rendered with crisp strand-level detail; visible pores between scales, layered plate edges, and precise specular reflections on each individual scale. The dragon exhales slowly in the cold air—warm breath condenses into swirling mist that curls around its snout and whiskers, with tiny ice crystals and micro-droplets visible in the light. Shallow depth of field: background mountain range heavily blurred into soft bokeh, while the dragon’s face remains tack-sharp. Cinematic natural sunlight, high contrast, ultra-detailed textures, hyper-real close-up, 8K resolution.

  \item A rugged, masculine man in an extreme close-up / tight close-up portrait, framing from forehead to upper chest, filling most of the frame. He has thick, voluminous hair and a dense, textured beard with clearly visible individual strands, subtle flyaways, and natural color gradients (dark base with slight warm highlights). Emphasize detailed skin texture: pores, light stubble transitions, fine wrinkles around the eyes, and realistic specular highlights. His eyes are deep and intense with sharp iris detail and reflective catchlights; strong jawline and defined cheek structure. Cinematic lighting with soft key light and gentle rim light that separates hair and beard from the background; shallow depth of field with smooth bokeh. Ultra-detailed textures, high realism, 8K resolution.
\end{enumerate}

\vspace{6pt}
\textbf{Figure 7 – Video Examples.}
\begin{enumerate}[leftmargin=1.2em,itemsep=2pt]
  \item A realistic close-up of an elderly man with gray hair and a thick gray beard, wearing a light-colored shirt. His head is slightly lowered. The camera zooms from full body to close-up, highlighting detailed facial wrinkles, skin texture, forehead lines, eye bags, and beard strands. High resolution, cinematic lighting, sharp details.
\end{enumerate}

\vspace{6pt}
\textbf{Figure 8 – Video Examples.}
\begin{enumerate}[leftmargin=1.2em,itemsep=2pt]
  \item An ultra-detailed, macro shot of an ancient, leather-bound alchemist's tome resting on a dark oak table. The focus is on the ornate cover and a heavy brass clasp. The leather is cracked, deep mahogany brown with visible grain, scuffs, and gold-leaf embossed occult symbols that are partially worn away. A heavy, tarnished brass latch shows intricate engravings and a subtle green patina in the crevices. Resting on the book is a pair of wire-rimmed spectacles with slightly dusty, scratched glass lenses reflecting a flickering candle flame. The background is a dim, moody study with blurred scrolls and glass vials. Warm, low-key candlelight creates deep shadows and glints on the metallic edges. Cinematic still life, hyper-realistic, 8K resolution, sharp focus on the leather texture.

  \item A hyper-realistic, tight close-up of an elderly nomad's hands holding a handful of parched, cracked desert earth. The skin is extremely weathered, dark tan, featuring deep, intersecting wrinkles, age spots, and ingrained dust in every crease. Fingernails are thick, uneven, and stained by the elements. A heavy, crude silver ring with a raw, cracked turquoise stone sits on one finger, showing a battered, hand-hammered texture. The dry earth in the palms is breaking into intricate geometric shards. Harsh, high-contrast desert sunlight creates dramatic highlights on the skin's ridges and deep shadows in the wrinkles. Cinematic documentary style, ultra-detailed textures, 8K resolution, sharp focus.

  \item An ultra-detailed macro photograph of a single fallen maple leaf on damp forest soil in autumn. The leaf is a vibrant mix of deep red, orange, and gold, with intricate, dark vein structures branching out across its surface. Countless tiny, perfect spherical dew droplets cling to the leaf, magnifying the texture beneath and reflecting the soft, overcast forest light like miniature lenses. The leaf surface shows natural imperfections, decay spots, and a slightly rough, organic texture. The background is a soft, dark bokeh of moist earth, moss, and other blurred colorful leaves. Natural, diffused lighting emphasizes the wetness and intricate details. Cinematic macro photography, photorealistic, 8K resolution, extremely shallow depth of field, razor-sharp focus on the center droplets.

  \item An extreme macro photograph of a single Scarlet Macaw feather, illuminated by translucent backlighting. The image focuses on the intricate microstructure of the feather. Clearly visible are the central shaft (rachis) with its fibrous texture, and the countless individual barbs and barbules interlocking like a complex microscopic mesh. The colors transition vividly from fiery red to bright yellow and deep blue, with natural gradients and incredible saturation. Tiny dust particles and microscopic imperfections cling to the barbs, enhancing realism. The backlighting makes the feather appear semi-translucent, revealing the complex internal network. Soft, natural light, shallow depth of field with a completely blurred background. Cinematic nature photography, photorealistic, ultra-detailed textures, 8K resolution.
\end{enumerate}

\vspace{6pt}
\textbf{Figure 9 – Video Examples.}
\begin{enumerate}[leftmargin=1.2em,itemsep=2pt]
  \item A cinematic close-up shot of an ancient, majestic sea turtle crawling slowly across a sun-drenched sandy beach. The turtle's skin is exceptionally detailed, showing deep, heavy wrinkles and a thick, leathery texture that reveals its great age. Its massive carapace is weathered, featuring deep cracks, intricate scutes, and a matte finish from years of sun and salt. Wet sand clings to its heavy flippers, leaving a distinct, textured trail in the soft shore behind it. In the background, the vast ocean's edge gently laps against the coast under the warm, golden light of a setting sun. Hyper-realistic skin and shell textures, 8K resolution, sharp focus on the turtle's aged face, dreamy coastal atmosphere, deep bokeh.

  \item An extreme close-up, frontal view of a majestic bald eagle soaring directly toward the camera. The focus is on its incredibly dense and layered plumage; individual feather barbs and downy under-feathers are visible, creating a 'hairy' and voluminous appearance. The eagle's eyes are sharp and piercing, with realistic reflections. The yellow beak shows fine ridges and natural imperfections. Its massive wings are spread wide, showing the complex interlocking structure of the primary feathers. Bright, high-altitude sunlight illuminates the feathers, creating soft shadows and highlighting the immense detail. Cinematic wildlife photography, 8K, photorealistic, razor-sharp focus.
\end{enumerate}

\vspace{6pt}
\textbf{Figure 10 – Video Examples.}
\begin{enumerate}[leftmargin=1.2em,itemsep=2pt]
  \item A solitary lighthouse standing on a rocky outcrop in the middle of a vast ocean, viewed from a distance across the sea. Wide cinematic shot showing the lighthouse rising above the waves against the horizon. The lighthouse structure is highly detailed with weathered stone walls, subtle cracks, layered masonry texture, metal railings, and a glass lantern room reflecting light. Powerful beams of light sweep slowly across the ocean surface. The sea is richly textured with rolling waves, foam patterns, and shimmering reflections of sunlight on the water. Mist and sea spray drift gently in the air, catching the light and creating a soft atmospheric glow around the lighthouse. The sky is expansive with layered clouds illuminated by warm sunset light. Cinematic composition, ultra-detailed textures, realistic ocean surface, volumetric lighting, high dynamic range, 8K resolution.

\end{enumerate}

\vspace{6pt}
\textbf{Figure 11 – Video Examples.}
\begin{enumerate}[leftmargin=1.2em,itemsep=2pt]
  \item A single white owl perched on a frost-covered pine branch in a snowy forest at dawn. Feathers rendered with crisp micro-detail, visible barbs and soft down, gentle breath mist in the cold air. Light snowfall drifting past the lens, pale blue fog weaving between dark pine trunks. The owl blinks slowly and turns its head with calm precision. Cinematic close-up with creamy bokeh, natural cold lighting, ultra-detailed texture, 8K, shallow depth of field.
  
  \item A powerful white tiger snarling in a snowy forest at dawn. Extreme close-up framing on the tiger’s face, filling most of the frame. Emphasize thick winter fur with layered strands and micro-texture, tiny snow crystals stuck to whiskers and brow fur, and subtle frost on the nose. The tiger’s lips curl back to reveal wet reflective teeth, detailed enamel edges, moist gums, and a few saliva threads glinting in the cold light. Warm breath condenses into swirling mist around the muzzle. Eyes are deep and intense with sharp iris patterns and reflective catchlights. Background is a heavily blurred snow-covered tree line with soft bokeh highlights. Ultra-detailed textures, cinematic winter lighting, high realism, 8K resolution.
  
  \item A hyper-realistic close-up of a female cybernetic pilot with advanced neural implants. Medium-close shot focusing on her temple and eye. The implants are made of brushed titanium and carbon fiber, featuring tiny glowing blue fiber-optic filaments, microscopic serial numbers, and realistic metallic oxidation. Her skin shows natural pores, fine peach fuzz, and subtle sweat beads around the interface. One eye is a sophisticated prosthetic with a multi-layered iris and tiny mechanical shutters. Soft ambient blue light from the cockpit displays reflects off her skin and metallic parts, creating a high-tech atmosphere. Cinematic sci-fi lighting, ultra-detailed textures, 8K resolution, sharp focus, physically based rendering.
  
  \item A close-up shot of a weathered leather satchel's brass buckle. The leather is aged and dark brown, showing a complex network of cracks, scuffs, and a rich, tactile grain. The heavy brass buckle is tarnished with a greenish patina in the corners and fine scratches from years of use. Thick, waxed linen stitching is visible, showing individual fibers and slight fraying. The lighting is warm and low-angled, emphasizing the 3D relief of the leather texture and the metallic sheen of the brass. Cinematic still life, 8K resolution, photorealistic materials, ultra-detailed, sharp focus.
  
  \item A hyper-realistic, top-down macro shot of a ceramic cup of espresso. The focus is on the rich, hazelnut-colored crema surface. The crema features a complex, marbled texture with tiny carbon dioxide bubbles and dark tiger-mottle patterns. A single, perfect silver spoon rests on the saucer, showing a mirror-polished surface with fine micro-scratches and a distorted reflection of the cafe environment. The ceramic cup has a subtle, tactile glaze with microscopic imperfections and a steam-fogged rim. Soft, directional light from a nearby window emphasizes the oily sheen and velvety texture of the coffee. Cinematic product photography, ultra-detailed, 8K resolution, sharp focus on the bubbles.

  \item An extreme macro photograph of a bumblebee perched on a deep purple lavender sprig. The focus is on the bee's thorax and head. Individual golden and black hairs (setae) are clearly visible, heavily dusted with microscopic, bright yellow pollen grains. Its compound eye shows a complex hexagonal lattice structure reflecting the sky. The wings are captured in a static, semi-translucent state, revealing a delicate network of dark veins and a shimmering, iridescent surface. Tiny droplets of morning dew cling to the bee's fuzz. Natural, soft morning sunlight creates a gentle rim light around the bee. Cinematic nature photography, hyper-realistic, 8K resolution, extremely shallow depth of field.

  \item An extreme macro shot of a human eye, focusing on the intricate patterns of the iris. The iris features complex, multi-layered radial textures in deep amber and forest green. The pupil is sharp and black. A tiny, detailed reflection of a sunlit window is visible on the cornea’s moist surface. Individual eyelashes are clearly defined with subtle clumps and natural curves. The sclera shows microscopic capillaries and a wet, glassy sheen. Natural skin texture around the eyelid reveals fine lines, pores, and microscopic hairs. Soft, directional light creates realistic depth and highlights. Cinematic macro photography, 8K resolution, ultra-detailed, hyper-realistic materials, razor-sharp focus.

  \item A close-up shot of a vintage 1950s camera lens. The focus is on the front glass element and the aperture rings. The glass shows multi-colored anti-reflective coatings with deep purple and amber hues. Tiny, realistic dust particles and faint cleaning marks are visible on the surface. The metallic lens barrel features knurled textures, engraved white and red focal length numbers with slight paint chipping. Oily, overlapping metal aperture blades are visible inside the lens. Soft studio lighting creates elegant curved highlights on the glass and metallic edges. Cinematic product photography, hyper-realistic textures, 8K resolution, sharp focus, industrial design.

\end{enumerate}

\vspace{6pt}
\textbf{Figure 12 – Video Examples.}
\begin{enumerate}[leftmargin=1.2em,itemsep=2pt]
  \item A peaceful coastal landscape with a tall lighthouse standing on a rocky cliff above the ocean, shown in a slightly closer cinematic view. Foreground rocks and waves are more visible while the lighthouse remains the main focus of the scene. Gentle waves crash against the rocks below while seagulls fly slowly across the sky. The lighthouse beam rotates slowly casting soft light across the water. The sunset sky glows with warm orange and golden colors reflecting across the ocean surface. The camera slowly moves toward the lighthouse capturing the tranquil seaside atmosphere, cinematic lighting, natural environment, highly detailed, 4k.

  \item A peaceful coastal landscape with a tall lighthouse standing on a rocky cliff above the ocean. Gentle waves crash against the rocks while seagulls fly across the sky and the lighthouse beam slowly rotates. Transform the entire scene into a dramatic Vincent van Gogh inspired oil painting style. The sky becomes a swirling pattern of deep blues, yellows, and oranges painted with thick expressive brush strokes. The ocean waves are rendered with curved textured strokes and vibrant colors.The rocky cliff and beach rocks are painted with thick impasto oil paint, visible swirling brush strokes, and bold textured strokes that follow the contours of the rocks. The rocks should appear clearly hand-painted with layered oil paint, expressive curved strokes, and rich painterly textures consistent with the surrounding Van Gogh style. The lighthouse stands boldly with glowing light, creating the appearance of a vivid post-impressionist oil painting on canvas, cinematic composition, highly detailed, 4k.

  \item A peaceful alpine landscape with a snow-covered mountain rising behind a calm lake surrounded by dense pine forests, shown in a slightly closer cinematic view. Smooth rocks appear along the lake shore in the foreground, especially on the left side, partially extending into the shallow water while reflecting softly on the lake surface. The mountain remains the central focus of the scene, illuminated by soft golden sunset light. Gentle ripples move across the lake while clouds drift slowly across the sky. The reflection of the snowy mountain and forest is clearly visible in the calm water. The camera slowly moves forward toward the lake capturing the tranquil mountain atmosphere, cinematic lighting, natural environment, highly detailed, 4k.

  \item Transform the original video into a dramatic Vincent van Gogh inspired oil painting style while strictly preserving the exact scene structure and composition from the source video. Apply a strong Vincent van Gogh inspired oil painting style to every element in the scene to create a clear visual transformation. The sky becomes a dramatic swirling pattern of blues, soft violets, and warm golden tones painted with thick expressive brush strokes. The snow covered mountain is rendered with bold textured oil paint,and visible hand painted highlights that emphasize the contours of the snowy ridges. The lake water is painted with curved flowing brush strokes and vibrant reflective colors, with swirling patterns moving across the surface in the style of Van Gogh. The reflections of the mountain and sky in the lake appear as painterly distorted brush textures rather than realistic reflections. The pine trees surrounding the lake are painted with energetic textured strokes and rich deep greens with visible oil paint layering. The shoreline rocks and foreground stones are painted with thick impasto oil paint, visible curved brush strokes, and rich painterly textures consistent with the surrounding Van Gogh style.Post-impressionist oil painting style, thick impasto brush strokes, textured canvas appearance, dramatic lighting, highly detailed, cinematic, 8k.

\end{enumerate}

\vspace{6pt}
\textbf{Figure 13 – Video Examples.}
\begin{enumerate}[leftmargin=1.2em,itemsep=2pt]
  \item An extreme close-up of a slice of hot pepperoni pizza being lifted, showing long, thin strands of melted mozzarella cheese stretching and breaking in slow motion. Small beads of oil glisten on the surface of the spicy pepperoni. Steam rises gently from the golden-brown crust. Soft, warm ambient lighting emphasizes the gooey, elastic texture of the cheese. The camera tracks the upward movement of the slice. Balanced appetizing lighting, cinematic food commercial style, ultra-detailed textures, 8k.

  \item A freshly baked loaf of bread resting on a wooden kitchen table in the morning. Warm sunlight streams through a nearby window, illuminating the golden crust and subtle cracks on the surface. Gentle steam rises from the bread, suggesting it has just come out of the oven. Tiny crumbs sit around the loaf on the rustic table. The camera slowly moves closer, revealing the crisp texture and warm tones of the crust. Cozy home kitchen atmosphere, soft natural lighting, realistic food photography, ultra detailed texture, 8k.

  \item A colorful butterfly resting gently on a blooming flower in a bright garden. Soft sunlight illuminates the delicate wings showing intricate patterns and vibrant colors. The butterfly slowly opens and closes its wings while the flower sways slightly in the breeze. The camera moves slowly closer for a detailed macro view. Balanced natural lighting, vivid colors, macro nature photography, ultra detailed textures, 8k.

  \item A small countryside cottage sitting peacefully in the middle of a green meadow under a clear blue sky. The stone walls and wooden roof look warm and rustic, with soft sunlight illuminating the texture of the bricks. Gentle wind moves the grass around the house. The camera slowly moves forward toward the cottage, highlighting the charming details of the windows, chimney, and wooden door. Bright natural daylight, calm rural atmosphere, realistic architectural details, cinematic depth of field, 8k.

  \item A simple notebook lying open on a wooden desk in a quiet study room. Soft afternoon sunlight falls across the pages, revealing faint handwritten notes and the subtle texture of the paper. A gentle breeze from a nearby window slowly flips one corner of the page. The camera gradually moves closer, focusing on the delicate fibers of the paper and the warm wooden surface beneath. Calm workspace atmosphere, soft natural lighting, shallow depth of field, realistic lifestyle scene, 8k.

  \item A sleek futuristic delivery drone hovering steadily above a modern city rooftop during a clear afternoon. The drone's metallic body reflects soft daylight while its propellers spin smoothly. The skyline stretches into the distance with bright glass buildings reflecting the sun. The camera slowly circles around the hovering drone, showcasing its detailed mechanical design. Balanced daylight, sci-fi technology aesthetic, cinematic motion, ultra detailed, 8k.

  \item A golden retriever sitting calmly on a grassy field during a warm sunny afternoon. Gentle sunlight highlights the dog's soft golden fur while a light breeze moves the grass around it. The dog slowly turns its head and looks curiously toward the camera. The camera slowly moves closer capturing the detailed texture of the fur and bright eyes. Balanced natural lighting, cheerful outdoor atmosphere, ultra detailed animal portrait, cinematic shot, 8k.

  \item A glowing jellyfish drifting slowly through clear ocean water. Soft sunlight filters down from the surface creating gentle light rays around the jellyfish. Its translucent tentacles move gracefully with the ocean currents. The camera follows the jellyfish smoothly as it floats through the water. Balanced underwater lighting, calm marine atmosphere, ultra detailed translucent textures, cinematic underwater scene, 8k.

  \item A medium shot of a kind elderly man with a bald head, wearing a simple gray t-shirt, standing in a brightly lit plain room, neutral expression, 8k.

  \item A futuristic city street at night filled with tall skyscrapers and glowing billboards. Cars move slowly along wet streets reflecting colorful lights. Neon signs in different colors illuminate the buildings while people walk along the sidewalks under umbrellas. Light rain falls gently, creating reflections and soft atmospheric haze. The camera slowly moves forward through the street, capturing the vibrant nightlife of the futuristic city, cinematic lighting, ultra detailed, 8k.

  \item A wide cinematic shot of a lone camel walking slowly across golden sand dunes in a vast desert at sunset. The camel leaves soft footprints in the sand while warm orange sunlight casts long shadows across the dunes. The wind gently moves the sand, creating subtle ripples on the surface. In the background, distant mountains are barely visible through a soft haze. The sky is filled with warm colors, transitioning from deep orange to soft pink and purple as the sun sets. The scene feels calm, expansive, and peaceful, with realistic lighting and natural movement, ultra detailed, cinematic atmosphere, 8k.

  \item Two small brown rabbits sitting together in a garden, nibbling on green leaves, soft daylight, blurred natural background, simple textures, 8k.

  \item A single warm street lantern hanging on a brick wall in a quiet narrow alley at night. Light rain falls gently through the air while the golden lantern light illuminates the wet cobblestone street below. Small raindrops shimmer as they pass through the warm light. The surrounding alley fades softly into the misty distance. The camera slowly moves forward along the street toward the lantern, capturing the calm cinematic night atmosphere, realistic lighting, highly detailed environment, 8k.

  \item A simple porcelain teacup placed on a wooden table beside a window in the morning. Warm sunlight shines through the window illuminating the cup and creating soft shadows on the table. Thin steam slowly rises from the hot tea swirling gently in the air. The camera slowly moves closer to the cup capturing the peaceful morning atmosphere. Natural warm lighting, cozy lifestyle scene, cinematic composition, ultra detailed textures, 8k.

  \item Soft pink cherry blossom branches hanging over a calm blue pond on a peaceful spring morning. A few delicate petals slowly fall from the branches and land gently on the water surface, creating small circular ripples that spread outward. The reflection of the blossoms and sky shimmers softly in the water. The camera slowly moves forward across the water surface, capturing the tranquil spring atmosphere. Natural daylight, soft colors, peaceful environment, cinematic composition, highly detailed, 8k.

  \item A magical turtle resting on a moss-covered rock in the middle of a calm forest stream at sunset. The turtle's shell glows with colorful sparkling lights that shimmer like tiny stars. Small glowing particles float gently around the turtle while the water flows slowly around the rock. Soft golden sunlight reflects across the water surface. The camera slowly moves closer toward the glowing shell capturing the magical atmosphere, cinematic fantasy lighting, highly detailed textures, 8k.

\end{enumerate}

\vspace{6pt}
\textbf{Figure 14 – Video Examples.}
\begin{enumerate}[leftmargin=1.2em,itemsep=2pt]
  \item  A confident young woman standing on a balcony overlooking a glowing city skyline at sunset. Her long hair flows gently in the wind as warm golden sunlight illuminates her face. She wears an elegant black evening dress with delicate silver jewelry. In the background, skyscrapers gradually light up as the city transitions into night. The camera slowly circles around her while soft wind moves the fabric of her dress. Cinematic portrait, soft lighting, shallow depth of field, ultra realistic, 8k.

  \item A powerful warrior standing in the center of an ancient stone temple hall. Massive statues line the walls while flaming torches illuminate the chamber with warm firelight. The warrior wears detailed historical armor decorated with metal plates and leather straps, holding a heavy weapon at his side. Dramatic beams of sunlight shine down from openings in the temple ceiling, cutting through dust particles in the air. The atmosphere feels epic and cinematic, ancient architecture, heroic pose, ultra detailed textures, dramatic lighting, 8k.

  \item  An ultra-luxury hypercar parked inside a futuristic architectural showroom at night, sculpted aerodynamic body with flowing lines, mirror-polished metallic paint reflecting soft ambient light, seamless carbon fiber details, low and wide stance emphasizing power and elegance, minimalistic environment with smooth concrete walls and large glass panels, subtle reflections on the glossy floor, cinematic wide-angle composition, soft rim lighting highlighting the curves of the vehicle, high dynamic range, neutral color palette, ultra-realistic materials, extreme detail, sharp focus, 8k ultra-high resolution, premium automotive photography, luxury and sophistication atmosphere.

  \item  An artistic cinematic close-up of an elegant strawberry dessert, glossy cream layers, fresh strawberries with tiny water droplets, delicate sugar decorations, soft pastel color palette, natural window light, shallow depth of field, macro photography, ultra-detailed textures, realistic reflections, fine dessert styling, clean background, photorealistic, 8K, high-end food photography, masterpiece

  \item A grand ancient temple city standing beneath a stormy sky. Massive stone temples and towering pillars rise from the ground as golden lightning tears across swirling clouds above. The sky glows with intense energy as bolts of lightning strike the temple structures, illuminating intricate carvings and ancient architecture. The atmosphere is epic, mythological, and cinematic, with powerful light rays and storm clouds moving dramatically.

  \item  An epic cinematic mountain landscape at golden hour,dramatic clouds rolling through the valleys, a crystal-clear river reflecting warm sunlight, ancient pine trees in the foreground, ultra-realistic rock and snow textures, atmospheric perspective, cinematic color grading, soft haze, volumetric lighting, ultra-wide angle, high dynamic range, hyper-detailed, photorealism, 8K, masterpiece quality

  \item A peaceful macro nature scene,delicate purple flowers covered with sparkling morning dew. Tiny water droplets cling to the petals and reflect soft light. The background is filled with smooth green and blue bokeh circles, creating a dreamy shallow depth-of-field effect. The flowers appear fresh, luminous, and extremely detailed.The two scenes are separated by a clean horizontal boundary, creating a strong contrast between epic mythology and intimate natural beauty. Ultra detailed, cinematic lighting, high dynamic range, 8k.

  \item A magnificent steampunk flying machine floating above a historic European city at sunset. The enormous airship is built with intricate brass gears, copper pipes, rotating propellers and a glass observation dome. Below it lies an old city filled with gothic towers, cathedral rooftops and narrow streets. Warm golden sunset light bathes the entire city while mist drifts between buildings. The airship slowly glides through the sky, its mechanical parts glowing with warm metallic reflections. Cinematic atmosphere, ultra detailed steampunk machinery, volumetric light, epic scale, 8k.

  \item A sleek luxury sports car driving along a coastal highway at sunset. The golden sunlight reflects off the polished metallic body while waves crash against cliffs beside the road. The car accelerates smoothly as the camera follows from a low cinematic angle, capturing reflections and motion blur. The sky glows with vibrant orange and pink colors. High-end automotive commercial, ultra realistic reflections, cinematic motion, 8k.

  \item  A luxury mechanical wristwatch displayed on a rotating stand in a modern showroom. Soft studio lights highlight the polished metal case and the intricate moving gears visible through the transparent back. Reflections glide across the glass surface as the watch slowly rotates. The camera performs a smooth cinematic close-up, revealing the precision craftsmanship. Balanced studio lighting, elegant product showcase, ultra detailed macro shot, 8k.

  \item A clear mountain waterfall cascading over smooth rocks in a lush green valley. Sunlight filters through tall trees, creating moving beams of light across the flowing water. Mist rises gently from the waterfall while small birds fly across the scene. The camera slowly glides toward the waterfall, capturing the sparkling water droplets and vibrant plants nearby. Balanced daylight, peaceful nature scene, ultra detailed, cinematic landscape, 8k.

  \item A young deer standing in a peaceful forest clearing during early morning. Sunlight filters through leaves creating soft patterns of light on the grass. The deer looks around curiously while butterflies flutter nearby. Leaves move gently in the breeze as the camera slowly approaches the animal. Balanced natural lighting, calm wildlife scene, ultra detailed fur texture, cinematic nature shot, 8k.

  \item A skilled potter shaping a spinning clay pot on a pottery wheel inside a bright artisan workshop. Soft daylight streams through large windows, illuminating clay textures and tools scattered on the wooden table. The pot slowly takes shape as the wheel spins and tiny clay particles splash outward. The camera gently moves closer, capturing the detailed movement of the hands and the smooth surface of the forming vase. Balanced natural lighting, warm studio atmosphere, ultra detailed, cinematic, 8k.

  \item A crystal-clear alpine lake surrounded by snowy mountains under bright daylight. The calm water reflects the peaks and drifting clouds above. A gentle breeze creates small ripples across the surface while birds glide across the sky. The camera slowly rises upward revealing the breathtaking landscape. Balanced natural light, vivid colors, cinematic mountain scenery, ultra detailed, 8k.

  \item A beautiful fairy standing in a magical night forest surrounded by glowing flowers and sparkling particles. Her translucent wings emit soft luminous patterns resembling delicate veins of light. Moonlight filters through dense leaves creating a dreamy atmosphere. Tiny floating lights drift through the air while soft blue and green tones illuminate the forest. The fairy looks calm and mysterious, long flowing hair moving gently in the breeze. Cinematic fantasy lighting, magical atmosphere, ultra detailed, shallow depth of field, 8k.

  \item A futuristic jungle ecosystem filled with advanced machines and glowing plants. Massive robotic structures stand among lush tropical vegetation while beams of sunlight penetrate the forest canopy. Transparent energy plants glow softly on the forest floor, emitting blue bioluminescent light. A small futuristic vehicle moves through the dense jungle landscape. The scene combines nature and technology in harmony, with intricate mechanical details and vibrant vegetation. Cinematic lighting, sci-fi environment, ultra detailed, volumetric sunlight rays, 8k.

\end{enumerate}

\end{tcolorbox}

\end{document}